\PassOptionsToPackage{table,xcdraw}{xcolor}
\documentclass{article}

\PassOptionsToPackage{numbers, compress}{natbib}
\usepackage[final, dandb]{neurips_2025}

\usepackage{graphicx}
\usepackage[utf8]{inputenc} % allow utf-8 input
\usepackage[T1]{fontenc}    % use 8-bit T1 fonts
\usepackage{hyperref}       % hyperlinks

\usepackage{url}            % simple URL typesetting
\usepackage{booktabs}       % professional-quality tables
\usepackage{amsfonts}       % blackboard math symbols
\usepackage{nicefrac}       % compact symbols for 1/2, etc.
\usepackage{media9}
\usepackage{microtype}      % microtypography      % colors
\usepackage{amsmath}
\usepackage{tabularx}
\usepackage{pifont}
\usepackage[most]{tcolorbox}
\usepackage{subcaption}
\usepackage{makecell}
\usepackage{enumitem}
\def\eg{\emph{e.g.}}

\usepackage{amssymb}
\usepackage{pdfpages}
\usepackage{caption}
\usepackage{multirow}
\usepackage{multicol}

\definecolor{mypink}{RGB}{255,105,180}

\newcommand{\name}{\textsc{ExAct}}

\title{\textsc{ExAct}: A Video-Language Benchmark \\ for Expert Action Analysis}

\author{
Han Yi \quad Yulu Pan \quad Feihong He \quad Xinyu Liu \quad Benjamin Zhang \\
\textbf{Oluwatumininu Oguntola} \quad \textbf{Gedas Bertasius} \\
University of North Carolina at Chapel Hill \\
{\tt\small \{hanyi, yulupan, gedas\}@cs.unc.edu}, 
{\tt\small \{feihonhe, xinyu1, zhangben, oguntola\}@unc.edu}
}

\begin{document}

\maketitle

\begin{figure}[h]
\begin{center}
\vspace{-.25in}
\includegraphics[width=\linewidth]{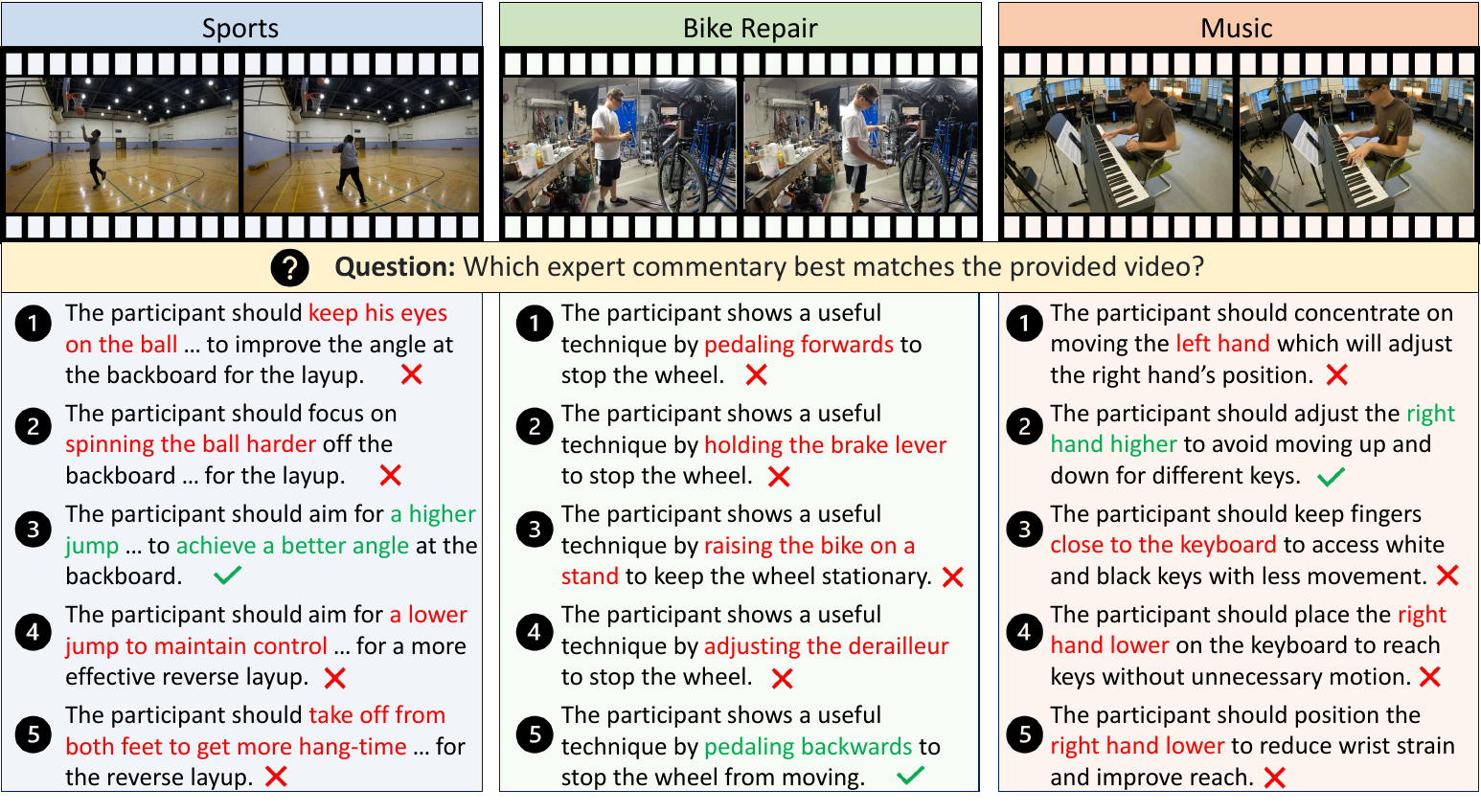}
\end{center}
\vspace{-.1in}
   \caption{ 
     An illustration of several multiple-choice question samples from our expert action analysis benchmark, \name. Here, we visualize samples from three domains of skilled activities (i.e., basketball, bike repair, and piano). The correct answers have a green checkmark next to them and were obtained using domain-expert/coach annotations. The phrases in green (correct) and red (incorrect) emphasize the subtle yet critical differences between ground-truth expert descriptions and incorrect candidate answers.
   }
\vspace{-.1in}
\label{fig:qa_samples}
\end{figure}

\begin{abstract}
  We present \name, a new video-language benchmark for expert-level understanding of skilled physical human activities. Our new benchmark contains 3{,}521 expert-curated video question-answer pairs spanning 11 physical activities in 6 domains: Sports, Bike Repair, Cooking, Health, Music, and Dance. \name~requires the correct answer to be selected from five carefully designed candidate options, thus necessitating a nuanced, fine-grained, expert-level understanding of physical human skills. Evaluating the recent state-of-the-art VLMs on \name~reveals a substantial performance gap relative to human expert performance. Specifically, the best-performing 
  Gemini 2.5 Pro model achieves only 55.35\% accuracy, well below the 82.02\% attained by trained human experts. We believe that \name~will be beneficial for developing and evaluating VLMs capable of precise understanding of human skills in various physical and procedural domains. Dataset and code are available at \href{https://texaser.github.io/exact_project_page/}{\textcolor{mypink}{https://texaser.github.io/exact\_project\_page/}}.

\end{abstract}

\section{Introduction}
\vspace{-0.2cm}

\begin{table*}[!b]
\vspace{-0.2cm}

\centering

{
\resizebox{\textwidth}{!}{
\begin{tabular}{lcccccc}
\toprule
\textbf{Dataset} & \textbf{Expert-level Knowledge} & \textbf{Free-form Language Annotations} & \textbf{MCQ Evaluation}
\\
\midrule

\multicolumn{4}{l}{\textit{Coarse Action Recognition Datasets}} \\

Kinetics-700~\cite{carreira2019short} & \textcolor{red}{\ding{55}} & \textcolor{red}{\ding{55}} & \textcolor{red}{\ding{55}}    \\

HowTo100M~\cite{miech2019howto100m} & \textcolor{red}{\ding{55}} & \textcolor{red}{\ding{55}} & \textcolor{red}{\ding{55}}  \\

UCF101~\cite{soomro2012ucf101} & \textcolor{red}{\ding{55}} & \textcolor{red}{\ding{55}} & \textcolor{red}{\ding{55}}  \\

HMDB~\cite{kuehne2011hmdb} & \textcolor{red}{\ding{55}} & \textcolor{red}{\ding{55}} & \textcolor{red}{\ding{55}} \\

Moments in Time~\cite{monfort2019moments} & \textcolor{red}{\ding{55}} & \textcolor{red}{\ding{55}} & \textcolor{red}{\ding{55}} \\

Hollywood~\cite{sigurdsson2016hollywood} & \textcolor{red}{\ding{55}} & \textcolor{green!50!black}{\ding{51}}  & \textcolor{red}{\ding{55}} \\

ActivityNet-QA~\cite{yu2019activitynet} & \textcolor{red}{\ding{55}} & \textcolor{red}{\ding{55}} & \textcolor{green!50!black}{\ding{51}} \\

\midrule

\multicolumn{4}{l}{\textit{Fine-grained Action Recognition Datasets}} \\

Something-SomethingV2~\cite{goyal2017something} & \textcolor{red}{\ding{55}} & \textcolor{red}{\ding{55}} & \textcolor{red}{\ding{55}}  \\

FineGym~\cite{shao2020finegym} & \textcolor{red}{\ding{55}} & \textcolor{red}{\ding{55}} & \textcolor{red}{\ding{55}}  \\

Multisports~\cite{li2021multisports} & \textcolor{red}{\ding{55}} & \textcolor{red}{\ding{55}} & \textcolor{red}{\ding{55}}  \\

TemporalBench~\cite{cai2024temporalbench} & \textcolor{red}{\ding{55}} & \textcolor{green!50!black}{\ding{51}} & \textcolor{green!50!black}{\ding{51}} \\
\midrule

\multicolumn{4}{l}{\textit{Video-Based Skill Assessment Datasets}} \\

JIGSAWS~\cite{ahmidi2017dataset} & \textcolor{green!50!black}{\ding{51}} & \textcolor{red}{\ding{55}} & \textcolor{red}{\ding{55}}  \\

Best~\cite{Doughty_2019_CVPR} & \textcolor{green!50!black}{\ding{51}} & \textcolor{red}{\ding{55}} & \textcolor{red}{\ding{55}}  \\

FineDiving~\cite{xu2022finediving} & \textcolor{green!50!black}{\ding{51}} & \textcolor{red}{\ding{55}} & \textcolor{red}{\ding{55}} \\

FP-Basket~\cite{bertasius2017baller} & \textcolor{green!50!black}{\ding{51}} & \textcolor{red}{\ding{55}} & \textcolor{red}{\ding{55}}\\

BASKET~\cite{pan2025basket} & \textcolor{green!50!black}{\ding{51}} & \textcolor{red}{\ding{55}} & \textcolor{red}{\ding{55}}  \\

Aifit~\cite{fieraru2021aifit} & \textcolor{green!50!black}{\ding{51}} & \textcolor{green!50!black}{\ding{51}} & \textcolor{red}{\ding{55}} \\

\midrule

\multicolumn{4}{l}{\textit{Skilled Activity Video-Language Datasets}} \\

VidDiffBench~\cite{burgess2025video} & \textcolor{green!50!black}{\ding{51}} & \textcolor{green!50!black}{\ding{51}} & \textcolor{red}{\ding{55}} \\

EgoExo-Fitness~\cite{li2024egoexo} & \textcolor{green!50!black}{\ding{51}} & \textcolor{green!50!black}{\ding{51}} & \textcolor{red}{\ding{55}} \\

EgoExolearn~\cite{huang2024egoexolearn} & \textcolor{green!50!black}{\ding{51}} & \textcolor{green!50!black}{\ding{51}} & \textcolor{red}{\ding{55}} \\

Ego-Exo4D~\cite{grauman2024ego} & \textcolor{green!50!black}{\ding{51}} & \textcolor{green!50!black}{\ding{51}} & \textcolor{red}{\ding{55}}\\
\midrule
\rowcolor{blue!10}
\textbf{\name~(Ours)} & \textcolor{green!50!black}{\ding{51}} & \textcolor{green!50!black}{\ding{51}} &\textcolor{green!50!black}{\ding{51}} \\
\bottomrule
\end{tabular}
}

    \caption{Compared to previous action recognition and skill assessment datasets, our proposed \name\ benchmark uniquely combines expert-level, free-form language annotations and a multiple-choice question (MCQ) evaluation format, making it an excellent resource for evaluating modern video-language models at expert-level understanding of skilled human activities.}
\label{tab:qa_dataset_comparison}

}
\end{table*}

Today, learning and perfecting a new physical skill requires a significant amount of time, practice, and often guidance from an expert coach/professional in that domain. Unfortunately, personalized coaching remains inaccessible to the majority due to prohibitively expensive costs and a lack of expert availability. Recent advances in AI have sparked interest in developing virtual AI assistants/coaches, particularly in the text domain~\cite{openai2024chatgpt, anthropic2023claude, team2024gemini}. 
However, text-based large language models (e.g., ChatGPT) are insufficient for learning new \textit{physical skills} as standard LLMs typically lack a nuanced understanding of physical human activities/skills. In contrast, rapidly improving vision-language models (VLMs) could serve as valuable tools for various physical skill learning applications. In particular, by recording a video of a skill demonstration and feeding it to a VLM, people could receive detailed, actionable feedback similar to that of expert human coaches.

Although recent progress has enabled modern VLMs to achieve impressive general image/video recognition capabilities~\cite{hurst2024gpt, team2024gemini, liu2023visual, bai2025qwen2, lin2023video, chen2024internvl}, existing studies reveal that such VLMs still struggle to understand fine-grained human activities~\cite{cai2024temporalbench, pan2025basket}, particularly activities that require expert-level knowledge~\cite{burgess2025video, huang2024egoexolearn, li2024egoexo}. This is primarily due to the inadequate underlying visual representations that 1) do not capture expert-level knowledge needed to generate feedback for physical skill learning, and 2) are unable to recognize subtle details in skilled human actions.

In addition to the limitations of existing VLMs, there is a notable lack of evaluation benchmarks tailored for expert-level understanding of skilled human activities. As shown in Table~\ref{tab:qa_dataset_comparison}, most existing datasets focus on coarse activity recognition~\cite{carreira2019short, miech2019howto100m, soomro2012ucf101, kuehne2011hmdb, monfort2019moments, sigurdsson2016hollywood, yu2019activitynet}, which typically only require scene-level recognition rather than fine-grained understanding. While several fine-grained video recognition datasets exist~\cite{goyal2017something, shao2020finegym, li2021multisports, cai2024temporalbench}, the majority of them are aimed at generic (\eg, putting something into something) rather than expert-level action understanding of skilled human activities (\eg, keeping balance during a dance spin, playing the correct rhythm in a piano piece). Beyond action recognition, a number of video-based skill assessment benchmarks have been recently developed~\cite{ahmidi2017dataset, Doughty_2019_CVPR, xu2022finediving, bertasius2017baller, pan2025basket, fieraru2021aifit}. Most of such skill assessment datasets focus on predicting scalar/categorical performance scores to quantify execution quality. Additionally, these datasets often lack open-ended language annotations, which can provide a rich and intuitive medium to convey skill-specific feedback~\cite{burgess2025video} and can capture subtle errors, temporal nuances, and intent—details that scalar or categorical labels often miss. Finally, a few recent skilled activity video-language datasets~\cite{burgess2025video, li2024egoexo, huang2024egoexolearn, grauman2024ego} use experts to obtain free-form language descriptions akin to verbal coach feedback but do not provide rigorous evaluation benchmarks or tasks, making it difficult to assess how well modern video models can understand nuanced physical human skills.

\begin{figure}[t]
\begin{center}
\vspace{-.25in}
\includegraphics[width=\linewidth]{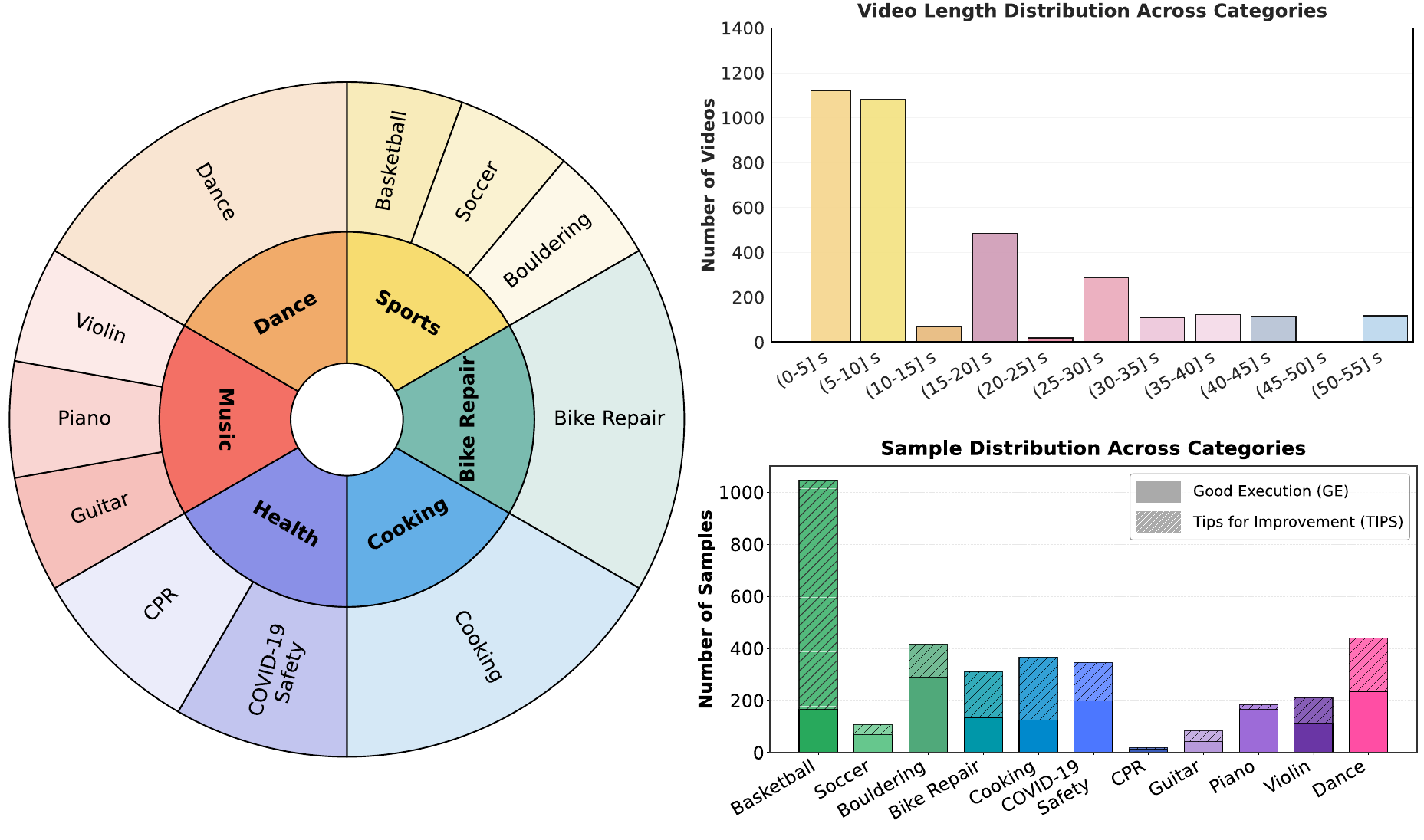}
\end{center}
\vspace{-.15in}
   \caption{\textbf{Left:} Our proposed \name~benchmark contains 11 skilled activity types spanning 6 broader physical domains: Sports, Music, Dance, Health, Cooking, and Bike Repair. \textbf{Top Right:} Distribution of video lengths across the dataset, showing that most clips fall within the 0–10 second range. \textbf{Bottom Right:} Sample distribution per activity, categorized by the expert feedback type: Good Execution (GE) and Tips for Improvement (TIPS). }
\vspace{-.32in}
\label{fig:distribution}
\end{figure}

To address these issues, we introduce \name, a video-language benchmark consisting of 3{,}521 video–question–answer (VQA) pairs designed to evaluate expert-level understanding of skilled physical human actions (sample questions shown in Figure~\ref{fig:qa_samples}). 
\name~covers 11 activities across 6 diverse physical domains: Sports (Basketball, Soccer, Bouldering), Bike Repair, Cooking, Health (COVID‑19 Safety, CPR), Music (Guitar, Piano, Violin), and Dance, as shown in Figure~\ref{fig:distribution}. To construct \name, we build on the fine-grained expert commentaries derived from the Ego-Exo4D dataset~\cite{grauman2024ego}. The original Ego-Exo4D expert commentaries have several crucial limitations: 1) the original expert commentaries are unstructured and lengthy, 2) they contain automatic speech recognition (ASR) errors, and 3) they include redundant or irrelevant content that is not directly tied to the observed actions. Furthermore, evaluating the quality of open-ended captions/descriptions in the original form of Ego-Exo4D expert commentaries is inherently difficult using standard language metrics such as CIDEr~\cite{vedantam2015cider}, BLEU~\cite{papineni2002bleu}, or ROUGE~\cite{lin2004rouge}, which do not accurately reflect the correctness or relevance of the instructional feedback in the context of skilled physical activities. To address these challenges, we first apply a structured annotation pipeline that rewrites the original commentaries into concise, self-contained feedback commentaries. Then, our proposed \name~benchmark evaluates expert action understanding as a multiple-choice question-answering task, which uses a well-defined metric of question-answering accuracy. This  multiple-choice
question-answering format eliminates the ambiguity and reproducibility issues associated with open-ended evaluation by providing clearly defined answers and controllable question difficulty through the design of distractor options.

We evaluate several state-of-the-art VLMs, including Gemini 2.5 Pro~\cite{comanici2025gemini},
GPT-4.1~\cite{hurst2024gpt}, GPT-4o~\cite{hurst2024gpt}, Gemini 1.5 Pro~\cite{team2024gemini}, LLaVA-Video~\cite{zhang2024video}, LLaVA-OneVision~\cite{li2024llava}, Qwen2.5-VL~\cite{bai2025qwen2}, VideoLLaMA~\cite{zhang2025videollama}, InternVL2.5~\cite{chen2024internvl}, and PerceptionLM~\cite{cho2025perceptionlm} on our new \name~benchmark. Our results reveal that compared to human experts, most modern VLMs achieve poor results on \name. In particular, Gemini 2.5 Pro, the best-performing model in our experiments, achieves 55.35\% accuracy. In comparison, non-expert humans achieve 61.86\% accuracy, while domain experts achieve 82.02\% accuracy. This substantial gap highlights the limitations of current VLMs in expert-level understanding of physical human skills. We hope that \name\ will serve as a rigorous evaluation benchmark for measuring expert-level understanding of skilled human actions, thus laying the foundation for AI systems that support enhanced human skill learning.

\section{Related Work}
% \vspace{-0.2cm}

\textbf{Vision-Language Models (VLMs).} Recent advances in Vision-Language Models (VLMs) have demonstrated impressive capabilities in understanding visual content. Models such as Gemini 2.5 Pro~\cite{comanici2025gemini},
GPT-4.1~\cite{hurst2024gpt}, GPT-4o~\cite{hurst2024gpt}, Gemini 1.5~\cite{team2024gemini}, LLaVA-OneVision~\cite{li2024llava}, Qwen2.5-VL~\cite{bai2025qwen2}, VideoLLaMA~\cite{zhang2025videollama}, and InternVL2.5~\cite{chen2024internvl} have achieved strong performance in tasks such as action recognition, video captioning, and visual question answering. Although these models show remarkable generalization, their outputs are often limited to high-level descriptions of the image/video content, rather than a detailed, fine-grained understanding of physical human actions and skills. More recent video-centric variants, such as LLaVA-Video~\citep{zhang2024video} and VideoLLaMA~\cite{zhang2025videollama}, attempt to extend static image capabilities to temporal inputs. However, these models still struggle to understand the fine-grained nuances of complex physical human activities, which are essential for human skill understanding. Specifically, most existing VLMs cannot assess how well a task is performed, nor articulate the strengths of the execution and the areas requiring improvement~\cite{burgess2025video, cai2024temporalbench, pan2025basket}. To address this gap, our proposed \name~provides a rigorous benchmark to evaluate expert-level understanding of physical human skills.

\textbf{Multimodal and VQA Benchmarks.} Diverse multimodal benchmarks have emerged to evaluate VLM performance, particularly in the video domain. ActivityNet-QA~\cite{yu2019activitynet} assesses the temporal reasoning of activities within longer videos via question-answering. NExT-QA~\cite{xiao2021next} focuses on compositional temporal reasoning. STAR~\cite{wu2024star} emphasizes situated reasoning. Benchmarks such as PerceptionTest~\citep{patraucean2023perception} probe the perceptual abilities of VLMs in various modalities. TemporalBench~\cite{cai2024temporalbench} targets fine-grained temporal understanding. MV-Bench~\cite{li2024mvbench} assesses temporal comprehension across 20 challenging tasks. EgoSchema~\cite{mangalam2023egoschema} focuses on egocentric human actions. MMWorld~\cite{he2024mmworld} aims to evaluate embodied agents in simulated environments. MLVU~\cite{zhou2024mlvu}, Video-MMMU~\cite{hu2025video}, MMVU~\cite{zhao2025mmvu}, Video-MMLU~\cite{song2025video}, and Video-MME~\cite{fu2024video} aim to evaluate modern MLLMs for complex video question answering tasks. Finally, SEED-Bench~\cite{li2023seed} and MMBench~\cite{liu2024mmbench} evaluate various multimodal abilities of MLLMs. However, most existing benchmarks focus on factual recall (e.g., “What color is the car?”), event-level understanding (e.g., “What action is being performed?”), or basic temporal reasoning (e.g., “What happened before this?”), and are not designed to capture the subtle nuances of skilled human actions. In contrast, our newly proposed \name~benchmark focuses explicitly on expert-level analysis of physical human skills.

\textbf{Video-based Skill Assessment Benchmarks.} Several recent works have focused on developing methods to assess human skills from video. Benchmarks such as MITDive~\cite{pirsiavash2014assessing}, UNLV-Dive~\cite{parmar2017learning}, MTL-AQA~\cite{parmar2019and}, and FineDiving~\cite{xu2022finediving} provide temporally segmented videos with action labels or action quality scoring for diving. FP-Basket~\cite{bertasius2017baller} and BASKET~\citep{pan2025basket}) focus on basketball, while LOGO~\cite{zhang2023logo}) provides human judgment scores for artistic swimming. Similar efforts also exist in other sports, including figure skating~\cite{pirsiavash2014assessing} and golf~\cite{mcnally2019golfdb}. Furthermore, datasets such as JIGSAWS~\cite{ahmidi2017dataset}, BEST~\cite{Doughty_2019_CVPR}, and EgoExoLearn~\cite{huang2024egoexolearn} focus on scenarios beyond sports, such as surgical tasks and daily activities. Most recently, the Ego-Exo4D~\cite{grauman2024ego} dataset introduces large-scale egocentric and exocentric video of skilled human activities. Ego-Exo4D includes spoken expert commentaries, offering a unique expert-level supervisory signal for understanding human skills. However, these expert commentaries are typically highly unstructured, noisy due to ASR errors, and often contain irrelevant information. Moreover, Ego-Exo4D does not provide a formal evaluation benchmark/task associated with such expert commentaries. In our work, we leverage such unstructured expert commentaries and construct a rigorous, expert-validated, and easy-to-evaluate \name~benchmark enabling evaluation of expert-level understanding of physical human actions/skills. \vspace{-0.3cm}
\begin{figure}[t]
\begin{center}
\includegraphics[width=\linewidth]{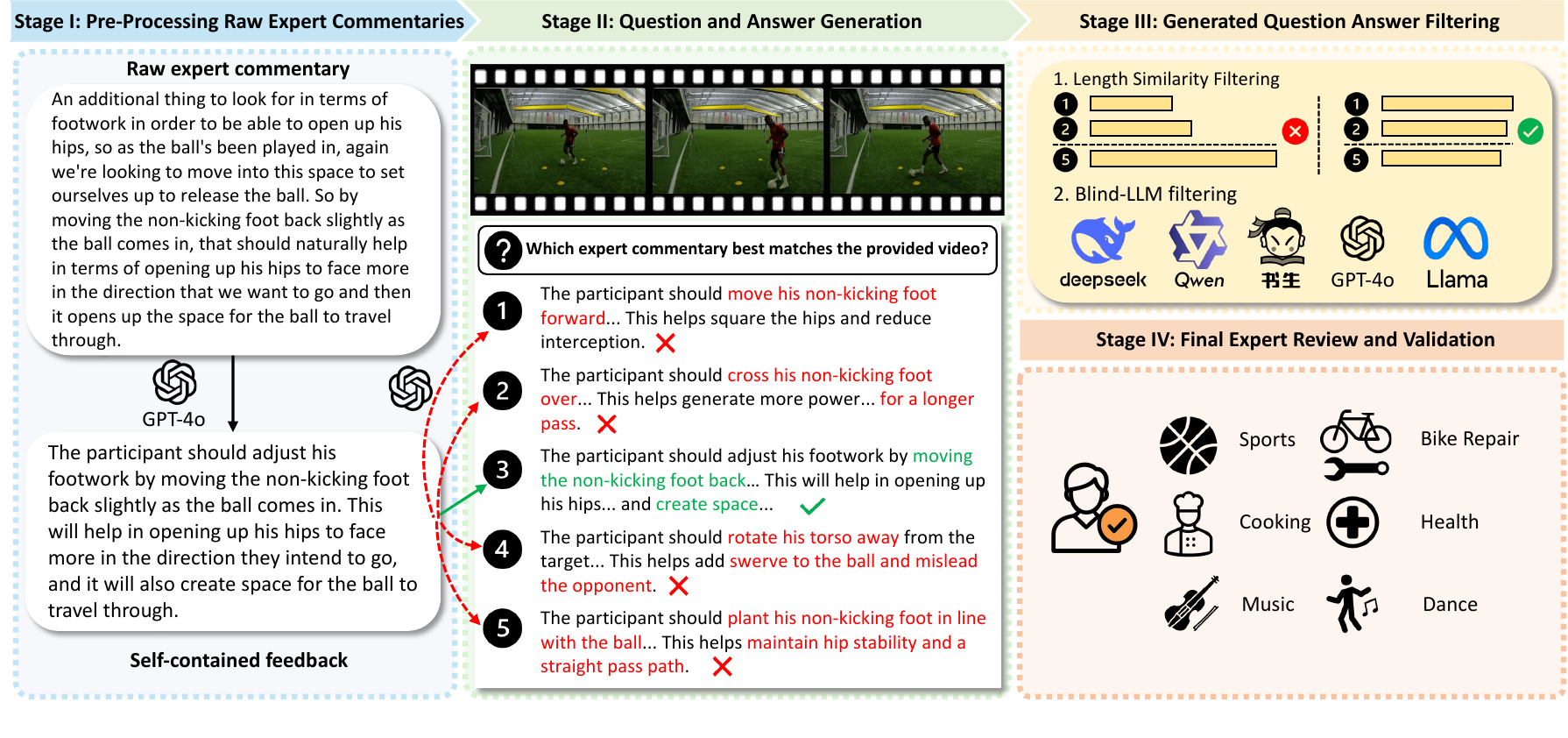}
\end{center}
\vspace{-0.3cm}
   \caption{Overview of our benchmark construction pipeline. In stage I, we pre-process raw expert commentaries using GPT-4o, correcting errors and segmenting them into concise, self-contained feedback commentaries. In Stage II, we construct multiple-choice QA pairs, each consisting of one correct expert commentary and four carefully generated distractors. The four red arrows indicate the LLM-generated distractors, while the green arrow represents the correct expert commentary. In Stage III, we filter out low-quality or biased samples using length-based heuristics and blind-LLMs. Finally, in Stage IV, domain experts review all QA pairs to ensure visual grounding and linguistic accuracy.\vspace{-0.3cm}
}
\label{fig:method}
\end{figure}

\section{\name~Benchmark Construction}

We construct \name\ using a four-stage pipeline. In stage I, we pre-process raw expert commentaries using GPT-4o, correcting  ASR errors and segmenting them into concise, self-contained feedback commentaries. In Stage II, we construct multiple-choice QA pairs, each consisting of one correct commentary and four distractors. In Stage III, we remove low-quality or biased samples through length filtering and blind-LLMs. Finally, in Stage IV, domain experts review each QA pair to ensure visual grounding and linguistic accuracy. Our benchmark construction pipeline is illustrated in Figure~\ref{fig:method}. \vspace{-0.2cm}

\subsection{Stage I: Pre-Processing Raw Expert Commentaries}
\label{sec:preprocessing}

The transcribed commentaries from Ego-Exo4D are often lengthy, noisy, and unstructured. They may also include automatic speech recognition (ASR) errors, redundant phrases, and off-topic remarks, as illustrated in the upper-left example of Figure~\ref{fig:method}. 
In the first stage of our benchmark construction pipeline, we use an LLM to refine raw expert commentaries from Ego-Exo4D into a more compact and less noisy format. Specifically, to do this, we prompt GPT-4o to 1) correct transcription mistakes, 2) remove irrelevant or repetitive content, and 3) segment the cleaned text into concise, self-contained feedback commentaries. 
Each commentary is then also assigned to one of the two categories: \textit{good execution} or \textit{tips for improvement}, reflecting the two main ways in which experts deliver their feedback, i.e., by affirming what was done well and/or suggesting what can be improved. We then use such compact and structured commentaries to construct multiple-choice question-answer pairs as described next. The complete prompt templates are provided in the supplementary material. \vspace{-0.2cm}

\subsection{Stage II: Question and Answer Generation} 
\label{sec:qa}

After pre-processing raw expert commentaries in Stage I, we proceed with multiple-choice QA pair construction. Specifically, we treat the structured commentary from Stage I as a positive commentary and ask the LLM (e.g., GPT-4o) to generate four distractor commentaries that require fine-grained expert-level understanding to distinguish the correct answer from the incorrect ones. This leads to a 5-way multiple-choice QA setup. To create negative commentaries, we use the following strategies: 

For expert commentaries assigned to the \textit{good execution} category (see Subsection~\ref{sec:preprocessing}), we apply two strategies to generate negative commentaries:
\begin{itemize}[itemsep=1pt, topsep=2pt, leftmargin=12pt]
    \item \textbf{Action replacement:} A key action in the original commentary is substituted with a plausible but incorrect alternative (\eg, replacing ``\textit{... performs a three-point shot}'' with ``\textit{... performs a layup}'').
    \item \textbf{Absent-action insertion:} A new event or action is inserted that was never mentioned or shown (\eg, adding ``\textit{... tightens the brake lever ...}'' to ``\textit{... keeps the bike steady on a repair stand ...}'').
\end{itemize}

For commentaries in the \textit{tips for improvement} category, we apply four alternative strategies:
\begin{itemize}[itemsep=1pt, topsep=2pt, leftmargin=12pt]
    \item \textbf{Action misinterpretation:} Misinterpreting the mistake in an execution.  
    \textit{Example:} Replacing ``\textit{... elbow is too low ...}'' with ``\textit{... grip is too tight ...}''.

    \item \textbf{Incorrect technical reasoning:} Correctly identifying a flaw but providing an implausible or technically inaccurate explanation.  
    \textit{Example:} ``\textit{... doesn’t bend knees ... reduces jump height}'' vs.\ ``\textit{... doesn’t bend knees ... prevents ball spin}'' (i.e., knee movement affects jumping, not ball spin).

    \item \textbf{False cause--effect relationship:} Introducing a misleading causal link between the error and an unrelated factor.  
    \textit{Example:} ``\textit{... feet are misaligned, leading to an off-balance shot}'' vs.\ ``\textit{... doesn’t tuck in jersey, leading to an off-balance shot}''.

    \item \textbf{Ineffective suggestion:} Proposing a correction to the execution that does not address the  problem.  
    \textit{Example:} ``\textit{... doesn’t keep eyes on the ball ...}'' vs.\ ``\textit{... should stand closer to the baseline}''.
\end{itemize}

Albeit simple in format, our questions effectively probe multi-faceted reasoning abilities—spanning temporal, causal, spatial, and domain-specific understanding—beyond mere visual recognition (see supplementary material~\ref{sec:reasoning} for examples).

\subsection{Stage III: Generated Question Answer Filtering}
\label{sec:qa filtering}

After constructing initial QA samples (Subsection~\ref{sec:qa}), we apply several additional filtering steps to ensure high-quality and unbiased samples. Prior studies~\cite{cai2024matryoshka, yue2024mmmu} have shown that large language models (LLMs) can often exploit subtle statistical or stylistic patterns in multiple-choice answers to correctly identify the ground-truth option without relying on visual input. Such language-driven shortcuts, often referred to as language-related bias, pose a serious threat to the integrity of robust evaluation. To mitigate such biases, we adopt two filtering strategies:

\noindent 1) \textbf{Length similarity:} Similar to human test-takers, LLMs may exploit surface-level cues such as answer length to eliminate implausible distractors. To mitigate this bias, we enforce a length similarity constraint: each distractor must be between 80\% and 120\% the length of the positive (i.e., correct) expert commentary, and the absolute word count difference must not exceed 8 words. QA samples that violate this constraint are excluded from the benchmark. Note that the 80–120\% length range and 8-word absolute difference threshold were determined empirically through iterative pilot studies with human annotators. %Annotators reported that noticeable length differences between answer choices often drew attention to superficial patterns over content.}

\noindent 2) \textbf{Blind-LLM filtering:} Inspired by findings from TemporalBench~\cite{cai2024temporalbench}, we observe that LLMs can sometimes identify the correct answer by detecting shared linguistic patterns, particularly when distractors are lightly edited variants of the ground truth. To avoid such language-driven biases, we present each QA sample consisting solely of the five textual options without any video to 5 state-of-the-art LLMs (GPT-4o and DeepSeek-VL-R1 (proprietary), Qwen2.5-72B, LLaMA3.3-70B, and InternLM2.5-20B (open-source, all ranked highly on multiple LLM leaderboards)). Each model is prompted with the question: \textit{``Which expert commentary best matches the provided video?''} If more than 20\% of the models (i.e., exceeding the random chance of selecting the correct answer in a 5-way setup) select the positive expert commentary, we consider the sample susceptible to language-only bias and remove it from the dataset.

\subsection{Stage IV: Final Expert Review and Validation}

In the final stage, we conduct a two-phase manual review to ensure that each QA question is clearly formulated and can be reliably answered through expert-level video analysis. Unlike the previous stages, which rely solely on textual inputs, this phase gives domain experts full access to the video alongside the five answer options. This setup allows for a comprehensive multimodal evaluation of answer correctness and distractor plausibility.

First, for each QA sample, %remove several
domain experts meticulously review the corresponding video clip and all five answer options. Their task is to select the option most consistent with the visual information presented in the video. After submitting their answer, the ground-truth label is revealed. The expert is then asked to verify three criteria: (1) whether the visual content of the video clip supports the correctness of the positive (i.e., correct) expert commentary; (2) whether any of the distractor commentaries also describe actions that are visible or valid within the video; and (3) whether all five candidate answer options are free from grammatical, logical, or instructional flaws. 
The sample will be removed if any of these criteria are not met. The second criterion is especially important because if a distractor candidate also describes something that appears in the video, the question becomes ill-posed due to multiple correct answers. For example, suppose that the positive expert commentary states, ``\textit{The participant plants their non-kicking foot beside the ball},'' while a distractor commentary states, ``\textit{The participant plants their foot behind the ball}.'' In this case, both candidate commentaries are visually observable, which means that the expert cannot answer it using a single answer. We remove all such samples to avoid ambiguity.

This verification process is conducted by 16 experts, each assigned to one of the 11 activity categories based on their area of expertise. Each sample is reviewed by at least one qualified expert to ensure consistency and domain relevance. For domains reviewed by two experts, we observed that experts reached agreement in over 90\% of cases—demonstrating strong consistency in their judgments. A screenshot of the annotation interface, along with additional implementation details, is provided in the supplementary material.
% \vspace{-0.2cm}

\section{Experimental Setup}
\label{sec:experiment setup}

\textbf{Evaluation Metrics.} We use standard question-answering accuracy as the primary evaluation metric. Our benchmark includes a total of 3{,}521 QA samples spanning 11 fine-grained physical skilled activities. For each activity, we compute the accuracy as the percentage of questions for which the model selects the correct answer. To summarize performance across the dataset, we report the average accuracy across all 11 activities. We additionally report per-domain accuracy (Sports, Bike Repair, Cooking, Health, Music, and Dance) to highlight domain-specific generalization.

\textbf{Baseline Models.}
To thoroughly assess the challenges posed by \name, we evaluate various state-of-the-art Video-Language Models (VLMs), including proprietary models: Gemini 2.5 Pro~\cite{comanici2025gemini},
GPT-4.1~\cite{hurst2024gpt}, GPT-4o~\cite{hurst2024gpt}, Gemini 1.5 Pro~\cite{team2024gemini}, and open-source models: LLaVA-Video~\cite{zhang2024video}, LLaVA-OneVision~\cite{li2024llava}, Qwen2.5-VL~\cite{bai2025qwen2}, VideoLLaMA~\cite{zhang2025videollama}, InternVL2.5~\cite{chen2024internvl}, and PerceptionLM~\cite{cho2025perceptionlm}. These models vary significantly in architecture, training corpus, and modality integration strategies, offering a broad and representative basis to evaluate expert-level feedback capabilities.

\textbf{Implementation Details.}
All model inferences are conducted using 4 NVIDIA Tesla H100 GPUs, each with 96 GB of memory. For fairness, we adopt uniformly sampling strategy and extract 32 frames per video clip for all models. Each frame is extracted at a resolution of 796~\(\times\)~448 and then resized internally according to the input resolution requirements of each model. Unless otherwise specified, we use the same prompt template and follow the official inference code provided by each model. We also ablate on several key hyperparameters in Subsection~\ref{sec:ablation}. Additional implementation details, including full prompt formulations, are provided in the supplementary material.

\section{Experimental Results}

\subsection{Main Results}

\begin{table*}[t]

\centering

{
\resizebox{\textwidth}{!}{
\begin{tabular}{lccccccc}
\toprule
\textbf{Model} & \textbf{Overall (\%)} & \multicolumn{6}{c}{\textbf{Results by Domain (\%)}} \\
\cmidrule(lr){3-8}
& & \textbf{Sports} & \textbf{Bike Repair} & \textbf{Cooking} & \textbf{Health} & \textbf{Music} & \textbf{Dance} \\
\midrule
Random Choice & 20.00 & 20.00 & 20.00 & 20.00 & 20.00 & 20.00 & 20.00\\
Human Non-Expert & 61.86 & 62.97 & 55.02 & 66.58 & 71.43 & 54.11 & 59.22\\
Human Expert & 82.02 & 82.09 & 81.23 & 80.27 & 87.09 & 80.21 & 81.55\\

\midrule
\rowcolor{blue!10}
\multicolumn{8}{l}{\textbf{Open-source VLMs}} \\
\midrule
PerceptionLM-8B~\cite{cho2025perceptionlm} & 24.65 & 24.22 & 28.16 & 25.75 & 22.53 & 22.95 & 26.42 \\
VideoLLaMA3-7B~\cite{zhang2025videollama} & 26.38 & 26.64 & 23.30 & 29.32 & 26.65 & 23.79 & 27.79\\
InternVL2.5-78B~\cite{chen2024internvl} & 33.48 & 31.93 & 36.57 & 33.70 & 37.91 & 32.00 & 34.62\\
LLaVA-OneVision-72B~\cite{li2024llava} & 35.44 & 33.65 & 43.04 & 33.42 & 35.44 & 30.53 & 43.51\\
Qwen2.5-VL-72B-Instruct~\cite{bai2025qwen2} & 35.67 & 35.62 & 37.86 & 33.97 & 36.26 & 32.63 & 38.50\\
LLaVA-Video-72B~\cite{zhang2024video} & 41.58 & 41.81 & 42.72 & 44.11 & 32.42 & 38.74 & 48.52\\

\midrule
\rowcolor{blue!20}
\multicolumn{8}{l}{\textbf{Proprietary VLMs}} \\
\midrule
Gemini 1.5 Pro~\cite{team2024gemini} & 43.91 & 42.83 & 52.10 & 51.78  & 41.21 & 41.89 & 39.86\\
GPT-4o~\cite{hurst2024gpt} & 44.70 & 43.47 & 52.75 & 46.30  & 53.30 & 33.89 & 46.70\\
GPT-4.1~\cite{hurst2024gpt} & 50.89 & 51.37 & 58.90 & 54.25  & 51.10 & 40.84 & 51.48\\
Gemini 2.5 Pro~\cite{comanici2025gemini} & \textbf{55.35} & \textbf{52.58} & \textbf{65.05} & \textbf{58.36} & \textbf{60.71} & \textbf{53.05} & \textbf{53.98}\\

\bottomrule
\end{tabular}
}

\caption{\name\ evaluation results (QA accuracy) across six diverse physical domains: Sports (Basketball, Soccer, Bouldering), Bike Repair,
Cooking, Health (COVID-19 safety, CPR), Music (Guitar, Piano, Violin), and Dance.  The results show a significant gap between the performance of modern Vision-Language Models (VLMs) and human experts, indicating that there is a significant room for improvement for future video-language models.
}
\label{tab:model_comparison}
}

\end{table*}

\begin{figure}[h]
\begin{center}

\includegraphics[width=\linewidth]{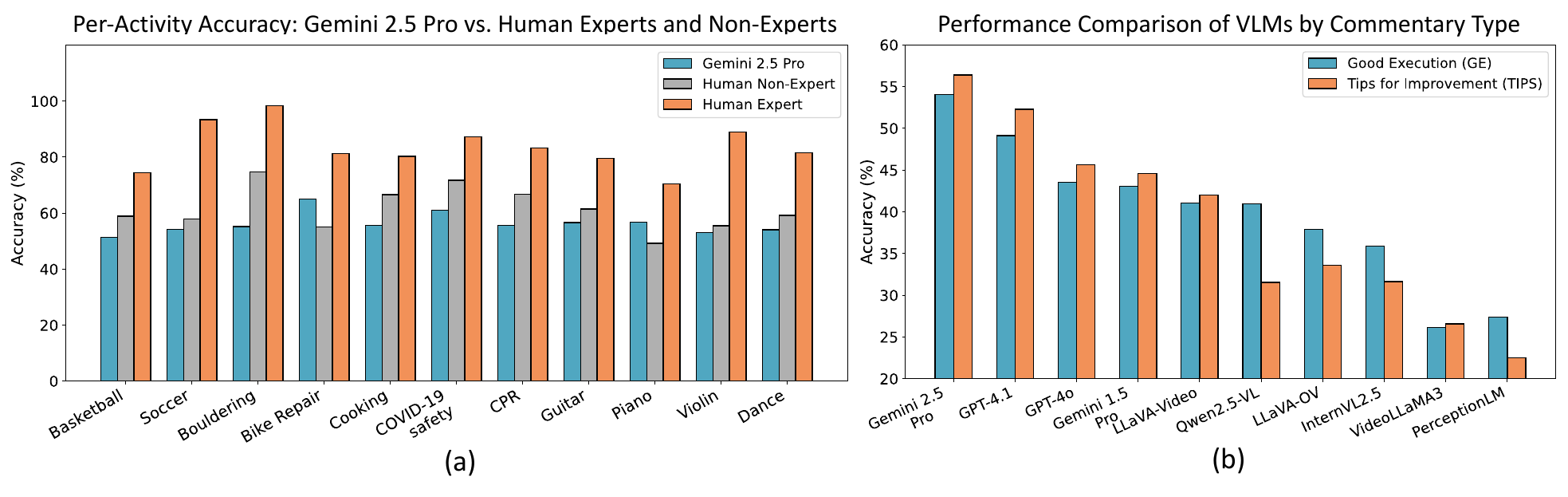}
\end{center}

   \caption{(a) Accuracy of the best-performing model (Gemini 2.5 Pro) compared to human experts and non-experts across domains. While Gemini 2.5 Pro achieves the highest accuracy among VLMs, it still falls short of human performance: experts consistently achieve over 80\% accuracy in most domains, with soccer and bouldering reaching up to 95\%. (b) Performance of VLMs on Good Execution (GE) and Tips for Improvement (TIPS) commentary categories. Models such as PerceptionLM, InternVL2.5, LLaVA-OneVision, and Qwen2.5-VL perform better on GE, while all proprietary models
   %like GPT-4o and Gemini 1.5 Pro 
   show stronger performance on TIPS. \vspace{-0.2cm}
}
\label{fig:model_feedback}
\end{figure}

% In Table~\ref{tab:model_comparison}, we report the performance of several state-of-the-art video-language models (VLMs) on our \name\ benchmark. Overall, all models perform poorly, with none exceeding 56\% accuracy compared with 20\% random-choice baseline. Among the evaluated models, Gemini 2.5 Pro achieves the highest overall accuracy at 55.35\%, followed closely by GPT-4.1 at 50.89\%. These two proprietary models outperform all others across all six domains. Notably, they achieve over 50\% accuracy in the domains of Bike Repair, Cooking, and Health. These domains tend to rely more heavily on procedural knowledge, such as understanding sequences of actions and planning the next steps. In contrast, domains like Sports, Music, and Dance involve highly specialized, fine-grained physical movements that require detailed visual perception and precise temporal understanding.

In Table~\ref{tab:model_comparison}, we report the performance of several state-of-the-art video-language models (VLMs) on our \name\ benchmark. Overall, none of the methods achieve accuracy exceeding 56\%. Among the evaluated models, Gemini 2.5 Pro achieves the highest overall accuracy at 55.35\%. This model outperforms all others across all six domains. Notably, it achieves over 60\% accuracy in the domains of Bike Repair and Health. These domains tend to rely more heavily on procedural knowledge, such as understanding sequences of actions and planning the next steps. In contrast, domains like Sports, Cooking, Music, and Dance involve highly specialized, fine-grained physical movements that require detailed visual perception and precise temporal understanding.

We also observe that proprietary models (e.g., GPT %GPT-4o
and Gemini) consistently outperform the best-performing open-source alternatives. This performance gap highlights the limitations of current public models in capturing the fine-grained, domain-specific reasoning required for expert-level skill understanding. Additionally, model size appears to be a significant factor in fine-grained video comprehension. Smaller models such as PerceptionLM-8B~\cite{cho2025perceptionlm} and VideoLLaMA3-7B~\cite{zhang2025videollama} achieve only 24.65\% and 26.38\% accuracy respectively—barely above random chance. In comparison, larger open-source models reach at least 33\% accuracy, suggesting that scale contributes to more effective representation and complex activity analysis.

\subsection{Human Performance Analysis}

To quantify the gap between human and model performance, we conduct a human evaluation involving two groups: \textbf{experts}, comprising trained coaches and professionals with significant domain expertise (i.e., $>10 \ years$), and \textbf{non-experts}, individuals without professional expertise in a given domain. As shown in Figure~\ref{fig:model_feedback} (a), experts achieve consistently high accuracy—often exceeding 80\% across most domains, with soccer and bouldering reaching around 95\%. Additionally, we observe that while non-experts perform considerably worse (60-70\% accuracy), they still %remove significantly 
surpass the best-performing VLM model (i.e., Gemini 2.5 Pro). These results highlight a substantial gap between human and VLM capabilities, emphasizing the challenges of \name{} and the limitations of current VLMs. 

\subsection{Performance by Expert Commentary Type}

Figure~\ref{fig:model_feedback} (b) shows the performance of various VLMs across two expert commentary categories: Good Execution (GE) and Tips for Improvement (TIPS). Models such as PerceptionLM~\cite{cho2025perceptionlm}, InternVL2.5~\cite{chen2024internvl}, LLaVA-OneVision~\cite{li2024llava}, and Qwen2.5-VL~\cite{bai2025qwen2} perform better on GE samples. However, they struggle with questions from the TIPS category, which require identifying subtle mistakes in skilled activity executions. In contrast, stronger models such as 
% GPT-4o~\cite{hurst2024gpt} and Gemini 1.5 Pro~\cite{team2024gemini}
Gemini 2.5 Pro~\cite{comanici2025gemini} and GPT-4.1~\cite{hurst2024gpt}
perform better on questions from the TIPS category, indicating a greater capacity for expert-level understanding of errors/mistakes in physical human activities.

\subsection{Ablation Studies}
\label{sec:ablation}

In Table~\ref{table:ablation}, we conduct ablation studies on four key design choices: (a) number of input frames, (b) impact of LLM size, (c) different prompt strategies, and (d) spatial video resolution. For (a)–(c), we use LLaVA-Video~\cite{zhang2024video}, the strongest-performing open-source baseline. We ablate the effect of spatial video resolution using Qwen2.5-VL, which supports native-resolution inputs.

\begin{table*}[t]
    \centering
    \renewcommand{\arraystretch}{1.25}
    \setlength\tabcolsep{8.0pt}
    {\fontsize{9}{9}\selectfont 
    
    \begin{minipage}[t]{0.32\linewidth}
        \vspace{0pt}
        \centering
        \resizebox{\textwidth}{!}{
            \begin{tabular}{l|c} 
                Num. Frames & Acc. (\%) \\
                \hline
                8 & 39.22 \\
                16 & 39.24 \\
                \rowcolor{blue!10}
                \textbf{32} & \textbf{41.58} \\
                64 & 39.73 \\
            \end{tabular}
        }
        \subcaption{
            \textbf{Number of Frames:} Using 32 input video frames leads to the best accuracy. 
        }
        \label{table:ablation-a}
    \end{minipage}
    \ 
    \begin{minipage}[t]{0.295\linewidth}
        \vspace{0pt}
        \centering
        \resizebox{\textwidth}{!}{
            \begin{tabular}{l|c} 
                Model Size & Acc. (\%) \\
                \hline
                7B & 27.07 \\
                \rowcolor{blue!10}
                \textbf{72B} & \textbf{41.58}  \\
            \end{tabular}
        }
        \vspace{0.32in}
        \subcaption{
            \textbf{Impact of LLM size:} Using larger LLMs leads to significantly better performance.
        }
        \label{table:ablation-b}
    \end{minipage}
    \
    \begin{minipage}[t]{0.355\linewidth}
        \vspace{0pt}
        \centering
        \resizebox{\textwidth}{!}{
            \begin{tabular}{l|c}
                Prompting & Acc. (\%) \\
                \hline
                w/o Time Info & 40.04 \\
                \rowcolor{blue!10}
                w/ Time Info & \textbf{41.58} \\
            \end{tabular}
        }
        \vspace{0.32in}
        \subcaption{
            \textbf{Prompting strategies:} Incorporating time information into the prompt improves accuracy.
        }
        \label{table:ablation-c}
    \end{minipage}
    \
    
    \caption{Ablation study on different design choices: number of input frames, impact of LLM size, different prompt strategies.}
    \vspace{-0.1cm}
    \label{table:ablation}
    \vspace{-0.2cm}
}
\end{table*}

\textbf{Number of Input Frames.}  Table~\ref{table:ablation-a} shows the effect of varying the number of input frames using a uniform sampling strategy. We observe a consistent improvement in performance as the number of frames increases, with the model achieving the highest accuracy at 32 frames. This suggests that incorporating more temporal context enhances the model’s ability to understand skilled human actions. We also observe that increasing the number of frames beyond 32 does not yield further gains, which may be because most VLMs are not optimized to process longer video sequences effectively.

\textbf{Impact of LLM Size.} In Table~\ref{table:ablation-b}, we use the same model architecture while varying the size of LLM in LLaVA-Video: 7B and 72B. As the size of LLM parameters increases, we observe a consistent improvement in accuracy. This trend highlights the critical role of the language model for capturing fine-grained video-language cues necessary for human skill analysis. %and reasoning in video-language tasks.

\textbf{Prompting Strategies.} In Table~\ref{table:ablation-c}, we analyze the impact of different prompting strategies on model performance. We first evaluate the role of including time information (\eg \textit{``The video is \{video\_time\}s long, and \{len(frames)\} uniformly sampled frames occur at \{frame\_time\}.''} in the prompt. Removing the time information leads to a performance degradation of 1.54\%, indicating that the understanding of time plays a meaningful role in the model’s analysis of human skill.

\textbf{Spatial Video Resolution.} We investigate how different spatial input resolutions affect the performance of Qwen2.5-VL by evaluating three settings: the original resolution of 796$\times$448 (35.67\% accuracy), a 1.5$\times$ downsampled version at 531$\times$299 (37.40\% accuracy), and a 2$\times$ downsampled version at 398$\times$224 (35.03\% accuracy). Interestingly, the model achieves the highest accuracy at the mid-level resolution of 531$\times$299. We hypothesize that although higher resolutions provide more visual detail, they may lead to performance degradation due to increased token length and a mismatch with the lower-resolution inputs commonly seen during pretraining.
\section{Conclusion}
\label{sec:conclusion}
We introduce \name, a new video-language benchmark designed to evaluate expert-level understanding of skilled human activities across a diverse set of physical and procedural domains. Our new benchmark uses fine-grained, expert-level, language annotations and a multiple-choice evaluation format to enable a rigorous evaluation of expert-level understanding of physical human skills. Our experiments reveal a significant gap between state-of-the-art VLMs and human experts' performance, indicating a significant room for future improvement in video-language model design. We believe that \name~will be pivotal in the development and evaluation of video language models capable of skilled human activity understanding.

\textbf{Limitations.} Although our benchmark spans multiple domains, it captures only a fraction of real-world activities. Additional tasks from underrepresented or specialized fields such as surgery or mechanical engineering may elicit different behaviors from current models and offer further insights. Moreover, certain domains (e.g., COVID-related tasks) may be time-sensitive or outdated, potentially affecting the relevance of some samples. Finally, while all participants consent to data usage, the inclusion of real-world videos containing identifiable human faces raises potential privacy concerns. The ethical implications surrounding the reuse and dissemination of such visual data warrant careful consideration and responsible handling.

\bibliography{references}
\bibliographystyle{plainnat} 

\newpage
\section*{NeurIPS Paper Checklist}

\begin{enumerate}

\item {\bf Claims}
    \item[] Question: Do the main claims made in the abstract and introduction accurately reflect the paper's contributions and scope?
    \item[] Answer: \answerYes{} % Replace by \answerYes{}, \answerNo{}, or \answerNA{}.
    \item[] Justification: The abstract and introduction clearly state the key contributions of the paper, including ``\name~will be beneficial for developing and evaluating VLMs capable of precise understanding of human skills in various physical and procedural domains.'' These claims are consistent with the technical content and experimental results presented in the paper. 

\item {\bf Limitations}
    \item[] Question: Does the paper discuss the limitations of the work performed by the authors?
    \item[] Answer: \answerYes{} % Replace by \answerYes{}, \answerNo{}, or \answerNA{}.
    \item[] Justification: We provide limitations in Section~\ref{sec:conclusion}.

\item {\bf Theory assumptions and proofs}
    \item[] Question: For each theoretical result, does the paper provide the full set of assumptions and a complete (and correct) proof?
    \item[] Answer: \answerNA{} % Replace by \answerYes{}, \answerNo{}, or \answerNA{}.
    \item[] Justification: The paper does not include any theoretical results.

    \item {\bf Experimental result reproducibility}
    \item[] Question: Does the paper fully disclose all the information needed to reproduce the main experimental results of the paper to the extent that it affects the main claims and/or conclusions of the paper (regardless of whether the code and data are provided or not)?
    \item[] Answer: \answerYes{}{} % Replace by \answerYes{}, \answerNo{}, or \answerNA{}.
    \item[] Justification: All necessary details to reproduce the main experimental results are provided in Section~\ref{sec:experiment setup}.

\item {\bf Open access to data and code}
    \item[] Question: Does the paper provide open access to the data and code, with sufficient instructions to faithfully reproduce the main experimental results, as described in supplemental material?
    \item[] Answer: \answerYes{} % Replace by \answerYes{}, \answerNo{}, or \answerNA{}.
    \item[] Justification: We provide open access to both the dataset and code used in our experiments. The dataset is available at: \url{https://huggingface.co/datasets/Alexhimself/ExAct}, and the evaluation code is available at: \url{https://github.com/Texaser/Exact}. Detailed instructions for reproducing the main experimental results are included in the repository and supplemental material.

\item {\bf Experimental setting/details}
    \item[] Question: Does the paper specify all the training and test details (e.g., data splits, hyperparameters, how they were chosen, type of optimizer, etc.) necessary to understand the results?
    \item[] Answer: \answerYes{} % Replace by \answerYes{}, \answerNo{}, or \answerNA{}.
    \item[] Justification: We provide the details in Section~\ref{sec:experiment setup}.

\item {\bf Experiment statistical significance}
    \item[] Question: Does the paper report error bars suitably and correctly defined or other appropriate information about the statistical significance of the experiments?
    \item[] Answer: \answerNo{} % Replace by \answerYes{}, \answerNo{}, or \answerNA{}.
    \item[] Justification: We do not report error bars. All results come from a single deterministic inference pass with pre-trained models—no model training or other stochastic procedures are involved—so repeated runs would yield identical numbers, while retraining the large models to obtain multiple seeds would be prohibitively expensive.

\item {\bf Experiments compute resources}
    \item[] Question: For each experiment, does the paper provide sufficient information on the computer resources (type of compute workers, memory, time of execution) needed to reproduce the experiments?
    \item[] Answer: \answerYes{} % Replace by \answerYes{}, \answerNo{}, or \answerNA{}.
    \item[] Justification: We provide the details in Section~\ref{sec:experiment setup}.
    
\item {\bf Code of ethics}
    \item[] Question: Does the research conducted in the paper conform, in every respect, with the NeurIPS Code of Ethics \url{https://neurips.cc/public/EthicsGuidelines}?
    \item[] Answer: \answerYes{} % Replace by \answerYes{}, \answerNo{}, or \answerNA{}.
    \item[] Justification: Our research adheres fully to the NeurIPS Code of Ethics. We ensured responsible use of data, preserved participant anonymity where applicable, and followed ethical practices in the collection, use, and reporting of all results. No known risks, harms, or ethical violations were introduced by this work.

\item {\bf Broader impacts}
    \item[] Question: Does the paper discuss both potential positive societal impacts and negative societal impacts of the work performed?
    \item[] Answer:  \answerYes{} % Replace by \answerYes{}, \answerNo{}, or \answerNA{}.
    \item[] Justification: The paper discusses both potential positive and negative societal impacts. It highlights the benefits of the proposed benchmark for evaluating expert-level understanding and supporting applications in education and skill training. In the limitations section, we also acknowledge potential negative societal impacts. Although all participants consented to data usage, the inclusion of real-world videos containing identifiable human faces may still pose privacy concerns.
    
\item {\bf Safeguards}
    \item[] Question: Does the paper describe safeguards that have been put in place for responsible release of data or models that have a high risk for misuse (e.g., pretrained language models, image generators, or scraped datasets)?
    \item[] Answer: \answerNA{} % Replace by \answerYes{}, \answerNo{}, or \answerNA{}.
    \item[] Justification: The benchmark does not involve the release of models or datasets with a high risk of misuse. All data sources are curated from publicly available, non-sensitive content with appropriate usage rights, and the benchmark is intended for academic research under a permissive license.

\item {\bf Licenses for existing assets}
    \item[] Question: Are the creators or original owners of assets (e.g., code, data, models), used in the paper, properly credited and are the license and terms of use explicitly mentioned and properly respected?
    \item[] Answer: \answerYes{} % Replace by \answerYes{}, \answerNo{}, or \answerNA{}.
    \item[] Justification: All external assets used in this work—including datasets, models, and code—are properly cited in the paper. Their respective licenses and terms of use (e.g., MIT, CC-BY, or custom academic licenses) have been reviewed and respected. We ensure that all assets are used in accordance with their intended academic and research use permissions.

\item {\bf New assets}
    \item[] Question: Are new assets introduced in the paper well documented and is the documentation provided alongside the assets?
    \item[] Answer: \answerYes{} % Replace by \answerYes{}, \answerNo{}, or \answerNA{}.
    \item[] Justification: The paper introduces a new benchmark, and we provide detailed documentation alongside the released assets. This includes task definitions, annotation guidelines, dataset statistics, data format specifications, and instructions for evaluation and usage. The documentation is provided alongside the released code and dataset at the uploaded URL.

\item {\bf Crowdsourcing and research with human subjects}
    \item[] Question: For crowdsourcing experiments and research with human subjects, does the paper include the full text of instructions given to participants and screenshots, if applicable, as well as details about compensation (if any)? 
    \item[] Answer: \answerYes{} % Replace by \answerYes{}, \answerNo{}, or \answerNA{}.
    \item[] Justification: As the paper involves research with human annotators, we provide the full text of the instructions given to participants, example screenshots of the annotation interface in the supplemental material.

\item {\bf Institutional review board (IRB) approvals or equivalent for research with human subjects}
    \item[] Question: Does the paper describe potential risks incurred by study participants, whether such risks were disclosed to the subjects, and whether Institutional Review Board (IRB) approvals (or an equivalent approval/review based on the requirements of your country or institution) were obtained?
    \item[] Answer: \answerYes{} % Replace by \answerYes{}, \answerNo{}, or \answerNA{}.
    \item[] Justification: The study involves human annotators evaluating non-anonymized video clips. While the videos may contain identifiable individuals, the annotation task itself poses no known risks to participants. All annotators were informed of the task content and compensation before participation. Based on our institutional guidelines, the nature of the task did not require formal IRB approval.

\item {\bf Declaration of LLM usage}
    \item[] Question: Does the paper describe the usage of LLMs if it is an important, original, or non-standard component of the core methods in this research? Note that if the LLM is used only for writing, editing, or formatting purposes and does not impact the core methodology, scientific rigorousness, or originality of the research, declaration is not required.
    %this research? 
    \item[] Answer: \answerYes{} % Replace by \answerYes{}, \answerNo{}, or \answerNA{}.
    \item[] Justification: Large language models (LLMs) are a core component of our evaluation methodology. Their usage is clearly documented in the main paper, including the specific models employed (e.g., GPT-4o, Gemini 1.5 Pro) and their roles within the evaluation pipeline.

\end{enumerate}

\newpage

\newpage
\appendix
\renewcommand{\thesection}{S\arabic{section}}
\setcounter{section}{0} 

\begin{center}
  {\Large \textbf{\name: A Video-Language Benchmark
\\for Expert Action Analysis}}\\[1.0em]
  {\large Supplementary Material}\\[1.0em]
\end{center}

Our supplementary materials include the following sections: Section \textcolor{red}{S1}: Full Prompts for Constructing \name, Section \textcolor{red}{S2}: Annotation Interface Details, Section \textcolor{red}{S3}: Qualitative Results, Section \textcolor{red}{S4}: Evaluated Reasoning Dimensions in \name, Section \textcolor{red}{S5}: Sample statistics after each filtering stage.

\section{Full Prompts for Constructing \name}
We provide the detailed prompts used for (1) pre-processing raw expert commentaries, (2) generating negative/distractor commentaries for \name, and (3) formatting inputs for VLMs.

\textbf{Prompt for Pre-Processing Raw Expert Commentaries.}
We present the prompts used for preprocessing raw expert commentaries into Good Execution and Tips for Improvement in Figure~\ref{fig:good execution prompt} and Figure~\ref{fig:tips prompt}, respectively. Corresponding examples of Good Execution and Tips for Improvement are shown in Figure~\ref{fig:good execution sample} and Figure~\ref{fig:tips sample}.

\begin{minipage}{\textwidth}
\begin{tcolorbox}[colback=cyan!10,        % light blue background
  colframe=white, % medium blue border
  boxrule=0.8pt,         % border thickness
  arc=2mm,               % rounded corners
  left=2mm, right=2mm, top=1mm, bottom=1mm, % internal padding
  enhanced,]

You are given a raw, transcribed expert commentary from a dataset describing a participant performing an \{activity\}. These transcriptions may contain automatic speech recognition (ASR) errors, redundant phrases, off-topic remarks, or unstructured language.

Your primary goal is to identify the good executions mentioned in the commentary and express them in a clean, concise, and coherent manner. Specifically:
\begin{enumerate}[nosep]
    \item Correct any transcription or grammatical errors.
    \item Remove irrelevant, repetitive, or filler content.
    \item Your goal is to determine the good executions mentioned in the commentary and write them in a coherent, concise manner.
    \item If there are no good executions, respond with: ``The expert mentions no good executions.''
\end{enumerate}

\end{tcolorbox}

\vspace{-1em}

\captionsetup{justification=raggedright,singlelinecheck=false}
\captionof{figure}{A prompt for pre-processing raw expert commentaries from the good execution (GE) category.
}
\label{fig:good execution prompt}
\end{minipage}

\begin{minipage}{\textwidth}
\begin{tcolorbox}[colback=yellow!20,        % light blue background
  colframe=white, % medium blue border
  boxrule=0.8pt,         % border thickness
  arc=2mm,               % rounded corners
  left=2mm, right=2mm, top=1mm, bottom=1mm, % internal padding
  enhanced,]

% --- Sample 1 ---
\textbf{Sample 1}\par
\vspace{0.5em}
\textbf{Input}\par
A grand great rhythm and flow as he crosses over. This is a great picture of going off the correct foot, which is his right foot, raising that left knee. He's got his eye and head up, total great body extension. He's got in great spatial awareness of where he is in relation to the basket and the rim. He's not too far under, he's not too far out.

\vspace{0.5em}

\textbf{Output}\par
The participant has a great rhythm and flow as he crosses over. He goes off of the correct foot, raises the correct knee, and has his head and eyes up. Overall, great body extension and spatial awareness under the rim.

\vspace{0.7em}

% --- Sample 2 ---
\textbf{Sample 2}\par
\vspace{0.5em}
\textbf{Input}\par
So here the dancer is jumping side to side, almost like a little short gallop. Really nice leg space. I definitely would recommend bending the knees a little bit more. And then just really like, yeah, I think if we bent the knees that would just kind of bring a better level and ease for him to jump side to side. Good job with the arms swaying back and forth on that part and then good eye connection to the audience facing front.

\vspace{0.5em}

\textbf{Output}\par
The participant demonstrates nice leg space during the jumps. Additionally, the participant nicely sways their arms and makes good eye contact with the audience.

\end{tcolorbox}

\vspace{-0.5em}

\captionsetup{justification=raggedright, singlelinecheck=false}
\captionof{figure}{Samples of converting raw expert commentaries into concise good execution (GE) commentaries.
}
\label{fig:good execution sample}
\end{minipage}

\begin{minipage}{\textwidth}
\begin{tcolorbox}[colback=cyan!10,        % light blue background
  colframe=white, % medium blue border
  boxrule=0.8pt,         % border thickness
  arc=2mm,               % rounded corners
  left=2mm, right=2mm, top=1mm, bottom=1mm, % internal padding
  enhanced,]
You are given a raw, transcribed expert commentary from a dataset describing a participant performing an \{activity\}. These transcriptions may contain automatic speech recognition (ASR) errors, redundant phrases, off-topic remarks, or unstructured language. 

Your primary goal is to identify the tips for improvement mentioned in the commentary and express them in a clean, concise, and coherent manner. Specifically:

\begin{enumerate}
    \item Correct any transcription or grammatical errors.
    \item Remove irrelevant, repetitive, or filler content.
    \item Your goal is to determine the tips for improvement mentioned in the commentary and write them in a coherent, concise manner.
    \item If there are no tips for improvement, respond with: ``The expert mentions no tips for improvement.''
\end{enumerate}
\end{tcolorbox}

\captionsetup{justification=raggedright,singlelinecheck=false}
\captionof{figure}{A prompt for pre-processing raw expert commentaries from the tips for improvement (TIPS) category.}

\label{fig:tips prompt}
\end{minipage}
\vspace{1.5em}

\begin{minipage}{\textwidth}
\begin{tcolorbox}[colback=yellow!20,        % light blue background
  colframe=white, % medium blue border
  boxrule=0.8pt,         % border thickness
  arc=2mm,               % rounded corners
  left=2mm, right=2mm, top=1mm, bottom=1mm, % internal padding
  enhanced,]

% --- Sample 1 ---
\textbf{Sample 1}\par
\vspace{0.5em}
\textbf{Input}\par
This is another example again of the person taking too heavy of a touch and then overextending his leg to be able to wrap around the ball. So actually you see now how he's contacting the ball. Instead of kind of catching it, he's actually having to almost like, his foot placement is much higher on the center of the ball here. He's actually coming like downward with the force in this direction when he's actually contacting it, which could actually potentially make the ball jump up when he hits it.

\vspace{0.5em}

\textbf{Output}\par
The participant should try to take a lighter touch and focus on not overextending their leg to wrap around the ball. The participant should avoid coming downward when contacting the ball because it could make the ball jump up when they hit it.

\vspace{1.2em}

% --- Sample 2 ---
\textbf{Sample 2}\par
\vspace{0.5em}
\textbf{Input}\par
This is a really excellent shift that the player had. If they wanted to be as expressive, they could change the weight as they're going up to come into the fingerboard a little bit more so it comes across as a more expressive slide.

\vspace{0.5em}

\textbf{Output}\par
The participant should change the weight as they go up to come into the fingerboard so it comes across as a more expressive slide.

\end{tcolorbox}

\vspace{0.5em}

\captionsetup{justification=raggedright, singlelinecheck=false}
\captionof{figure}{Samples of converting raw expert commentaries into concise tips for improvement (TIPS) commentaries.}
\label{fig:tips sample}
\end{minipage}

\newpage
\textbf{Prompt for Negative Commentary Generation.} Here we present the prompts used for negative/distractor commentary generation. For Good Execution (GE) commentaries, we adopt two strategies, as shown in Figure~\ref{fig:qa ge prompt}. For Tips for Improvement commentaries, we use four strategies, as shown in Figure~\ref{fig:qa tips prompt}.

\begin{minipage}{\textwidth}
\begin{tcolorbox}[colback=cyan!10,        % light blue background
  colframe=white, % medium blue border
  boxrule=0.8pt,         % border thickness
  arc=2mm,               % rounded corners
  left=2mm, right=2mm, top=1mm, bottom=1mm, % internal padding
  enhanced,]
You are a bouldering expert tasked with creating \textbf{wrong but plausible commentary} to train a video understanding model. You will be given high-expertise bouldering commentary, and your task is to generate \textbf{four wrong comments} based on that expert commentary. These wrong comments should be grounded in visible actions from the video and appear reasonable but must either provide an incorrect justification or directly misinterpret the actions. 
\\

\textbf{Requirements:}
\begin{itemize}[leftmargin=1.5em]
    \item All comments must be grounded in observable actions from the video and avoid references to non-visual elements.
    \item Match the length, detail, and complexity of the expert's original comments without obvious stylistic differences.
    \item The difference between correct and incorrect comments should lie in the reasoning or specific actions mentioned.
    \item Do not generate comments that might sometimes be true; ensure the actions are \textit{definitely} incorrect based on expert feedback.
    \item Avoid using negative adjective words such as ``improper,'' ``bad,'' ``not good,'' or ``not perfect.''
    \item Ensure the incorrect comments appear plausible, limiting each to 1--2 subtle errors.
    \item Vary the type of error across the four generated comments. Keep all comments logical and coherent.
\end{itemize}

\vspace{0.5em}
\textbf{Some techniques for creating wrong comments:}
\begin{itemize}[leftmargin=1.5em]
    \item \textbf{Action replacement:} A key action in the original commentary is substituted with a plausible but incorrect alternative.
    \item \textbf{Absent-action insertion:} A new event or action is inserted that was never mentioned or shown.
\end{itemize}

\vspace{1em}
\textbf{Output Format:}

\vspace{0.5em}
% \small
\textit{Good execution:} ``The participant demonstrates a good initiation of upward movement, properly preparing their legs to generate momentum upwards.''

\vspace{0.5em}
\textit{Wrong Comments:}
\begin{itemize}[leftmargin=1.5em]
    \item \textbf{Action replacement:} ``The participant demonstrates a good initiation of downward movement, correctly preparing their arms to generate momentum downwards.''
    
    \item \textbf{Action replacement:} ``The participant properly prepares their arms to generate momentum sideways, effectively aiding their lateral movement along the boulder.''
    
    \item \textbf{Absent-action insertion:} ``The participant demonstrates a clever strategy by swinging their body to the side before leaping to the next hold, avoiding direct upward movement.''
    
    \item \textbf{Absent-action insertion:} ``The participant initiates a powerful dyno by planting both feet on the foothold and lunging directly for the top of the wall, using agility over controlled movement.''
\end{itemize}

\vspace{0.5em}
\textit{The high-expertise comment is as follows:}
\end{tcolorbox}
\captionsetup{justification=raggedright, singlelinecheck=false}
\captionof{figure}{Prompt for negative/distractor commentary generation for good execution commentaries. We use a bouldering example for illustration.}
\label{fig:qa ge prompt}
\end{minipage}

\begin{minipage}{\textwidth}
\begin{tcolorbox}[colback=cyan!10,        % light blue background
  colframe=white, % medium blue border
  boxrule=0.8pt,         % border thickness
  arc=2mm,               % rounded corners
  left=2mm, right=2mm, top=1mm, bottom=1mm, % internal padding
  enhanced,]
You are a cooking expert tasked with creating \textbf{wrong but plausible commentary} to train a video understanding model. You will be given high-expertise cooking commentary, and your task is to generate \textbf{four wrong comments} based on that expert commentary. These wrong comments should be grounded in visible actions from the video and appear reasonable but must either provide an incorrect justification or directly misinterpret the actions. 
\\
% \vspace{1em}
\textbf{Requirements:}
\begin{itemize}[leftmargin=1.5em]
    \item All comments must be grounded in observable actions from the video and avoid references to non-visual elements.
    \item Match the length, detail, and complexity of the expert's original comments without obvious stylistic differences.
    \item The difference between correct and incorrect comments should lie in the reasoning or specific actions mentioned.
    \item Do not generate comments that might sometimes be true; ensure the actions are \textit{definitely} incorrect based on expert feedback.
    \item Avoid using negative adjective words such as ``improper,'' ``bad,'' ``not good,'' or ``not perfect.''
    \item Ensure the incorrect comments appear plausible, limiting each to 1--2 subtle errors.
    \item Vary the type of error across the four generated comments. Keep all comments logical and coherent.
\end{itemize}

\vspace{0.5em}
\textbf{Some techniques for creating wrong comments:}
\begin{itemize}[leftmargin=1.5em]
    \item \textbf{Action misinterpretation:} Misinterpreting the mistake in an execution. 
    \item \textbf{Incorrect technical reasoning:} Correctly identifying a flaw but providing an implausible or technically inaccurate explanation.
    \item \textbf{False cause–effect relationship:} Introducing a misleading causal link between the error and an unrelated factor.
    \item \textbf{Ineffective suggestion:} Proposing a correction to the execution that does not address the problem.
\end{itemize}

\vspace{1em}
\textbf{Output Format:}

\vspace{0.5em}
% \small
\textit{Tips for improvement:} ``The participant should add herbs, spices, and tea while waiting for the mixture to come to a simmer to improve efficiency and flavor infusion.''

\vspace{0.5em}
\textit{Wrong Comments:}
\begin{itemize}[leftmargin=1.5em]
    \item \textbf{Action misinterpretation:} ``The participant should wait until the mixture has finished simmering before adding herbs, spices, and tea, as this prevents any flavors from being cooked out of the ingredients.''
    
    \item \textbf{Incorrect technical reasoning:} ``The participant should add herbs, spices, and tea while the mixture is boiling vigorously, as the intense heat heightens the flavor of these ingredients.''
    
    \item \textbf{False cause–effect relationship:} ``The participant should add herbs, spices, and tea just after the mixture stops simmering, as this allows the flavors to cool simultaneously with the dish for balanced taste.''
    
    \item \textbf{Ineffective suggestion:} ``The participant should blend the herbs, spices, and tea into a smooth paste before adding them to the mixture after it comes to a simmer, ensuring a more uniform flavor throughout.''
\end{itemize}

\vspace{0.5em}
\textit{The high-expertise comment is as follows:}
\end{tcolorbox}

\captionsetup{justification=raggedright, singlelinecheck=false}
\captionof{figure}{Prompt for negative/distractor commentary generation from tips for improvement commentaries. We use a cooking example for illustration.}
\label{fig:qa tips prompt}

\end{minipage}

% \vspace{-.5in}
\newpage
\textbf{VLM Prompt for Processing \name.} Here we provide the template for the input prompt to LLaVA-Video. The prompts used for other models are very similar, only with a different video separation token. The default image token is the video separation input for LLaVA-Video. The scenario prompt briefly introduces the activity being performed by the participant (\eg, ``The participant is practicing basketball.''). We also include time-related instructions, such as the total video duration and the timestamps of uniformly sampled frames in the prompt to guide the VLMs. The prompt presents five candidate answer options labeled \textbf{Option 1} to \textbf{Option 5}, including one correct answer and four distractors.

\begin{minipage}{\textwidth}
\begin{tcolorbox}[
  colback=cyan!10,    % light background
  colframe=white,     % no visible border
  boxrule=0.8pt,
  arc=2mm,
  left=2mm, right=2mm, top=1mm, bottom=1mm,
  enhanced,
]

\small
\texttt{<DEFAULT\_IMAGE\_TOKEN>}

The video is \{\texttt{video\_time}\} seconds long, and \{\texttt{len(frames)}\} uniformly sampled frames occur at \{\texttt{frame\_time}\}.

\{\texttt{scenario\_prompt}\}

Below are different feedback statements about the person's performance in this video:

\textbf{Option 1.} \quad \textbf{Option 2.} \quad \textbf{Option 3.} \quad \textbf{Option 4.} \quad \textbf{Option 5.}

Based on what you observe in the video, which expert commentary best matches the provided video? \\
\textbf{Just respond with the option number (1--5) and nothing else.}

\end{tcolorbox}

\captionsetup{justification=raggedright, singlelinecheck=false}
\captionof{figure}{An input prompt to Vision-Language models (VLMs) for processing \name.}

\end{minipage}

\section{Annotation Interface Details}
\begin{figure}[h]
\begin{center}
\vspace{-.25in}
\includegraphics[width=\linewidth]{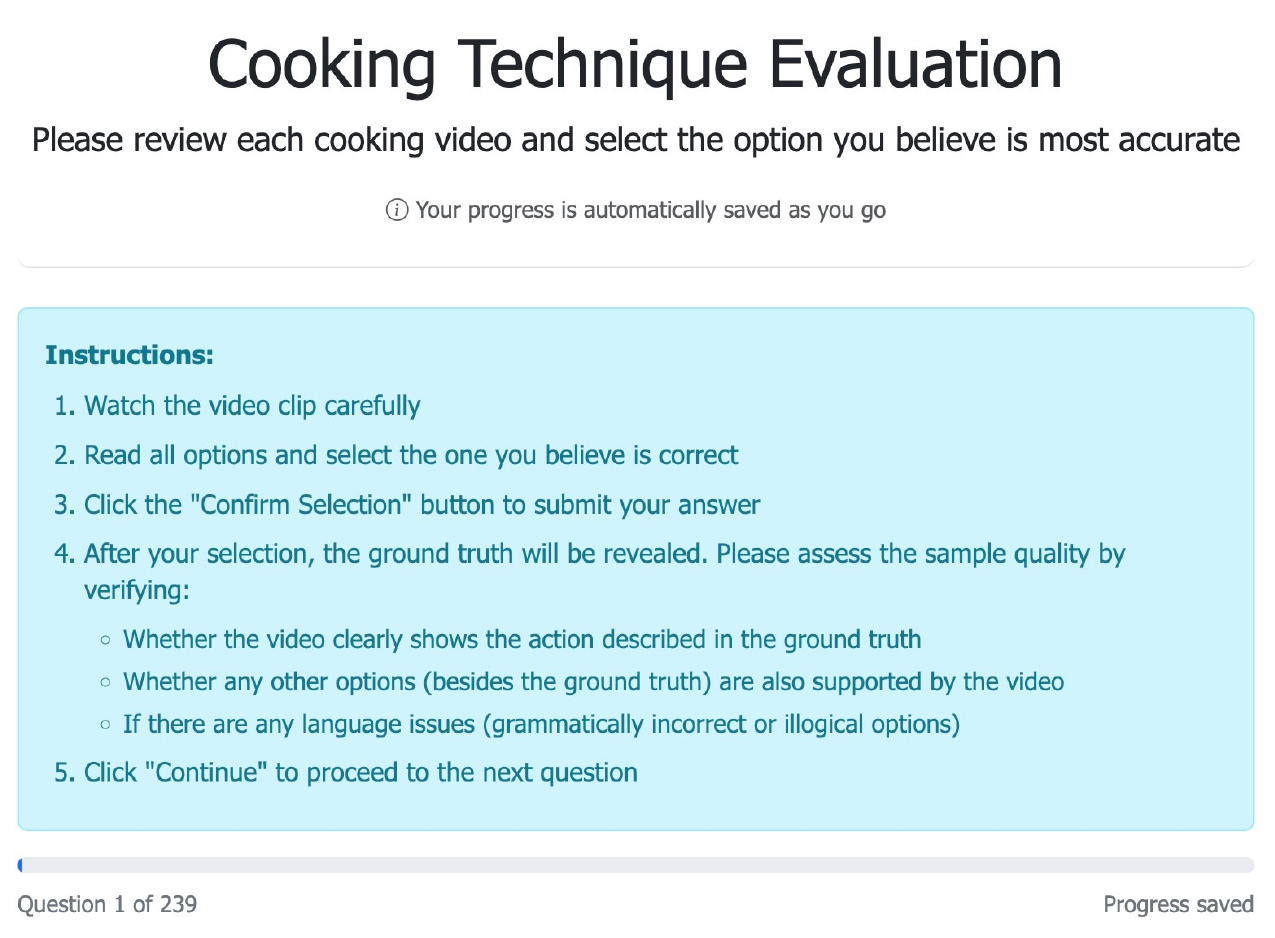}
\end{center}
% \vspace{-.15in}
   \caption{User instructions for the annotation platform interface.
}
\label{fig:instruction}
\end{figure}

\begin{figure}[h]
\begin{center}
\vspace{-.2in}
\includegraphics[width=\linewidth]{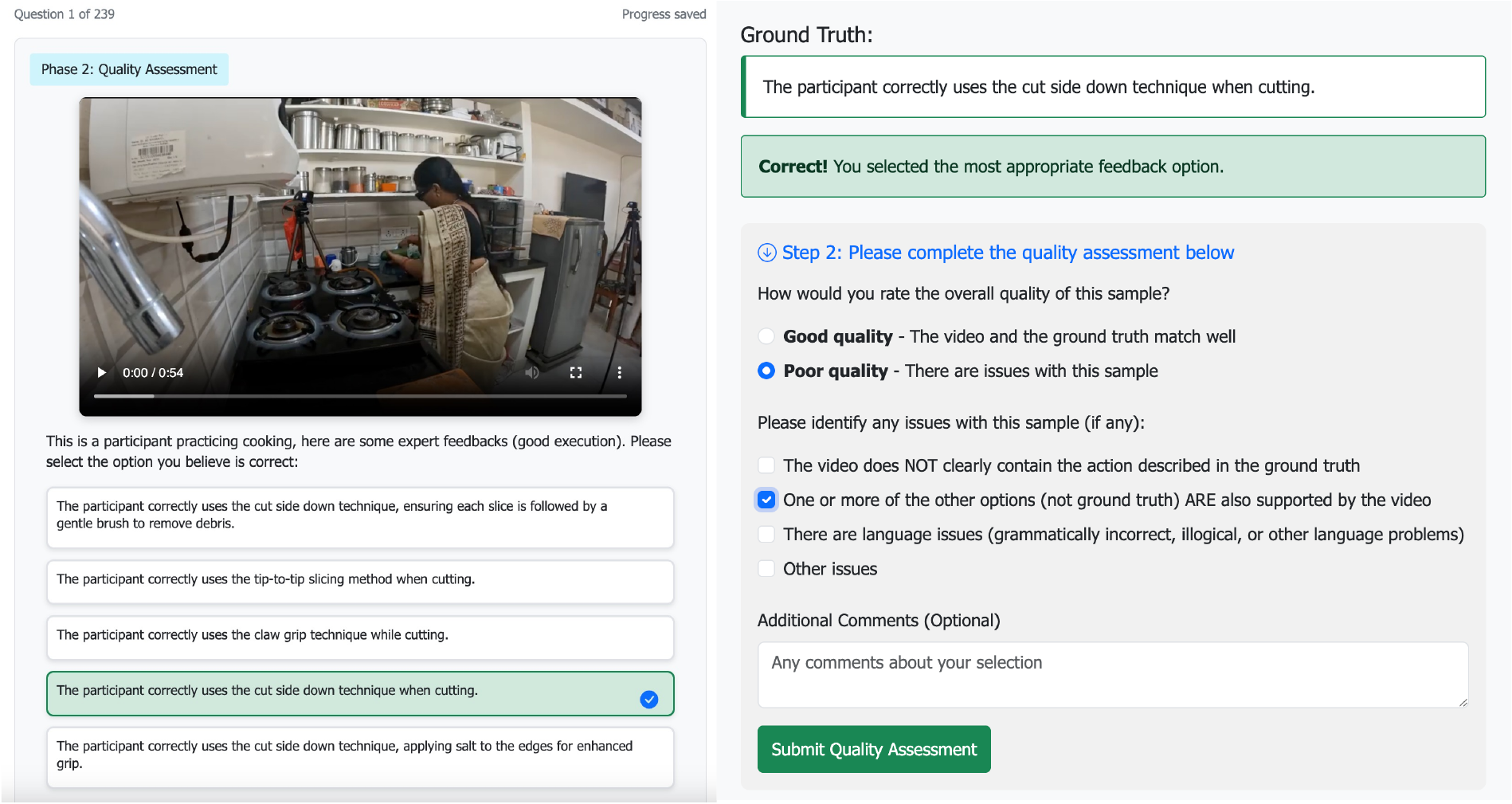}
\end{center}
\vspace{-.05in}
   \caption{Two-phase manual review interface on the annotation website.
}
\label{fig:review interface}
\end{figure}

In this section, we provide more details related to the annotation interface used for annotating \name. We develop a user-friendly web-based platform tailored for human annotators. A total of 16 experts participated in the annotation process, each with at least ten years of experience in their respective domains. The website is built using \texttt{github.io}, with \texttt{Formspree} used to collect submission data from annotators. All annotators are compensated at a rate of \$50 per hour. The interface begins with an introduction to the project, followed by detailed annotation guidelines. It supports saving progress, allowing annotators to complete their assigned tasks over multiple sessions rather than in a single sitting. Each expert is only assigned samples within their domain of expertise. Upon completing their assigned tasks, annotators submit their responses via the form. Figure~\ref{fig:instruction} and
Figure~\ref{fig:review interface} shows our data annotation interface.

\textbf{Guidelines for Annotators.} Please follow the instructions below when annotating:
\begin{enumerate}
    \item Carefully watch the video clip.
    \item Read all five options and select the one you believe is correct.
    \item Click the ``Confirm Selection'' button to submit your answer.
    \item After submitting your selection, the ground-truth answer will be revealed. Please then evaluate the sample based on the following criteria:
    \begin{enumerate}
        \item Does the video clearly support the action described in the ground-truth commentary?
        \item Do any of the other options also appear valid based on the video?
        \item Are there any language issues, such as grammatical errors or illogical phrasing, in any of the options?
    \end{enumerate}
    \item Click ``Continue'' to move on to the next sample.
\end{enumerate}

\newpage
\section{Qualitative Results}
\begin{figure}[t]
\begin{center}
\vspace{-.1in}
\includegraphics[width=\linewidth]{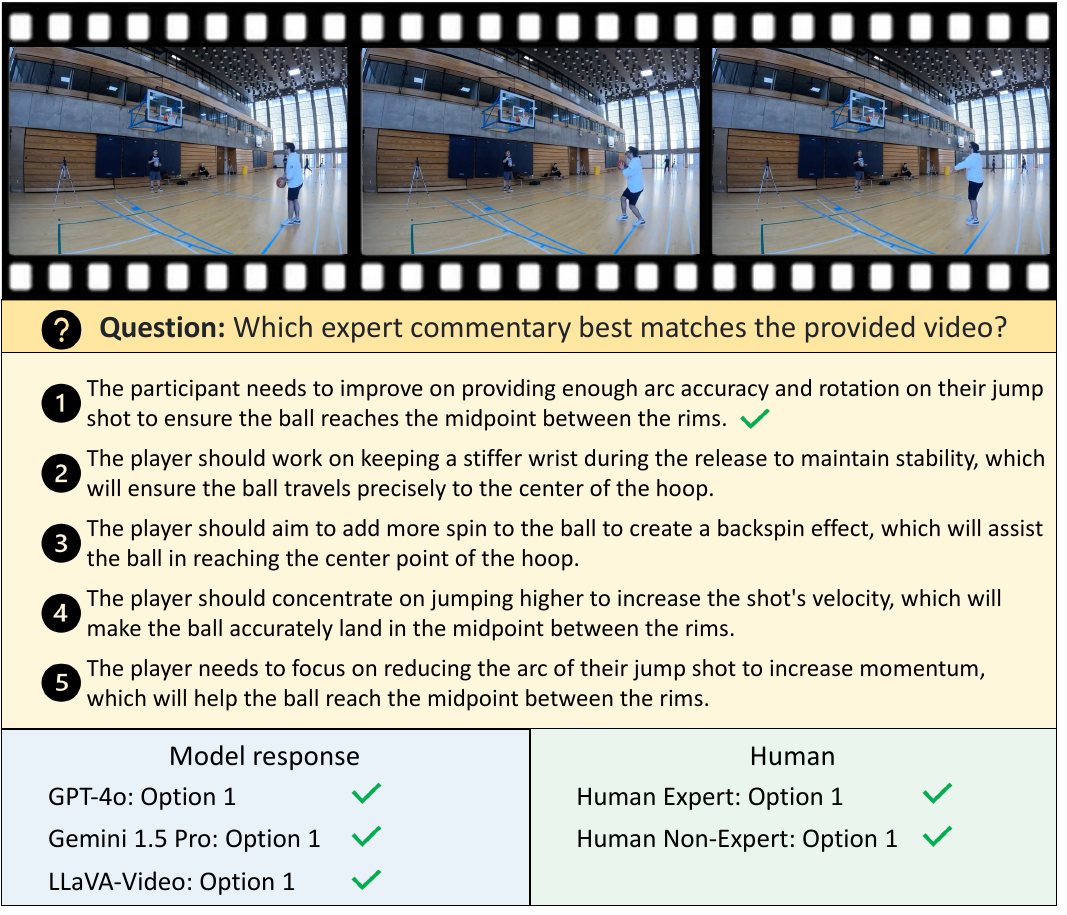}
\end{center}
\vspace{-.15in}
   \caption{Sample 1 (Basketball): All models, as well as both human experts and non-experts, select the correct answer.
}
\label{fig:sample 1}
\end{figure}

\begin{figure}[t]
\begin{center}

\includegraphics[width=\linewidth]{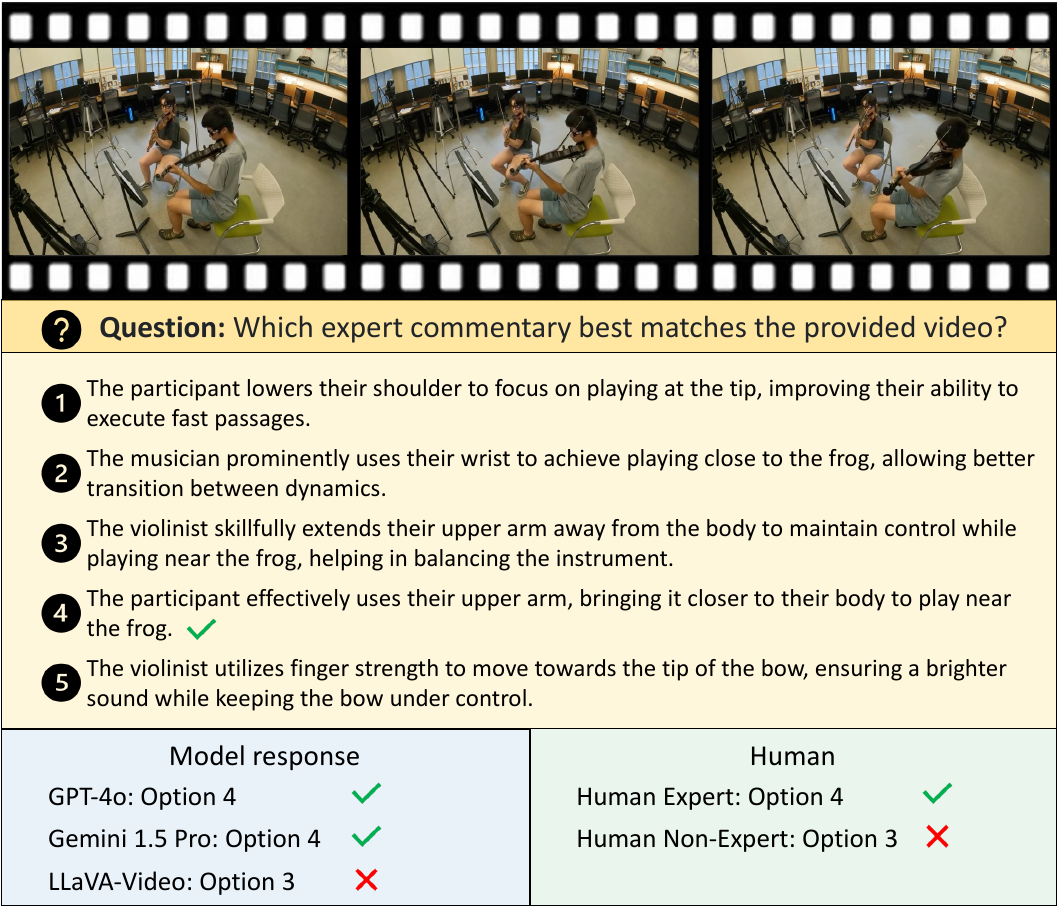}
\end{center}

   \caption{Sample 2 (Violin): Some models and human experts select the correct answer.
}
\label{fig:sample 2}
\end{figure}

\begin{figure}[t]
\begin{center}
% \vspace{-.1in}
\includegraphics[width=\linewidth]{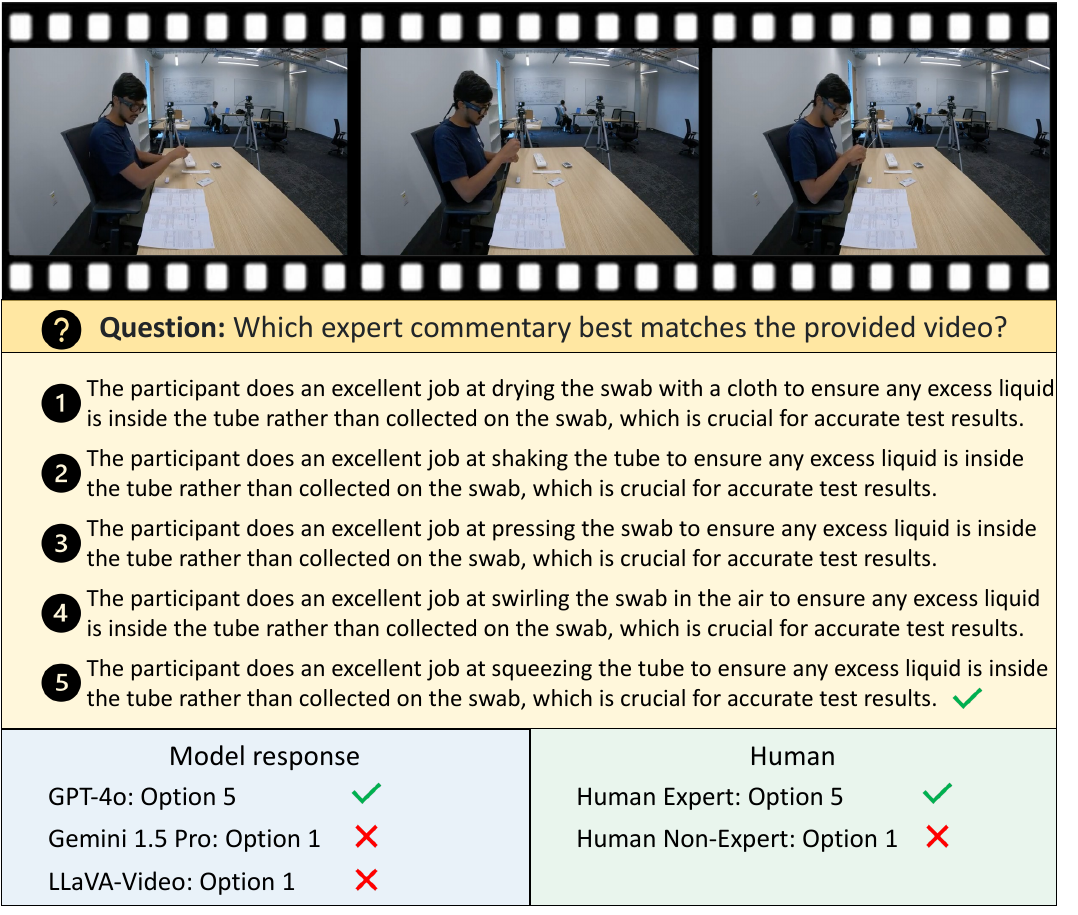}
\end{center}
% \vspace{-.15in}
   \caption{Sample 3 (COVID-19 Safety): Only GPT-4o and human experts select the correct answer.
}
\label{fig:sample 3}
\end{figure}

\begin{figure}[t]
\begin{center}
% \vspace{-.1in}
\includegraphics[width=\linewidth]{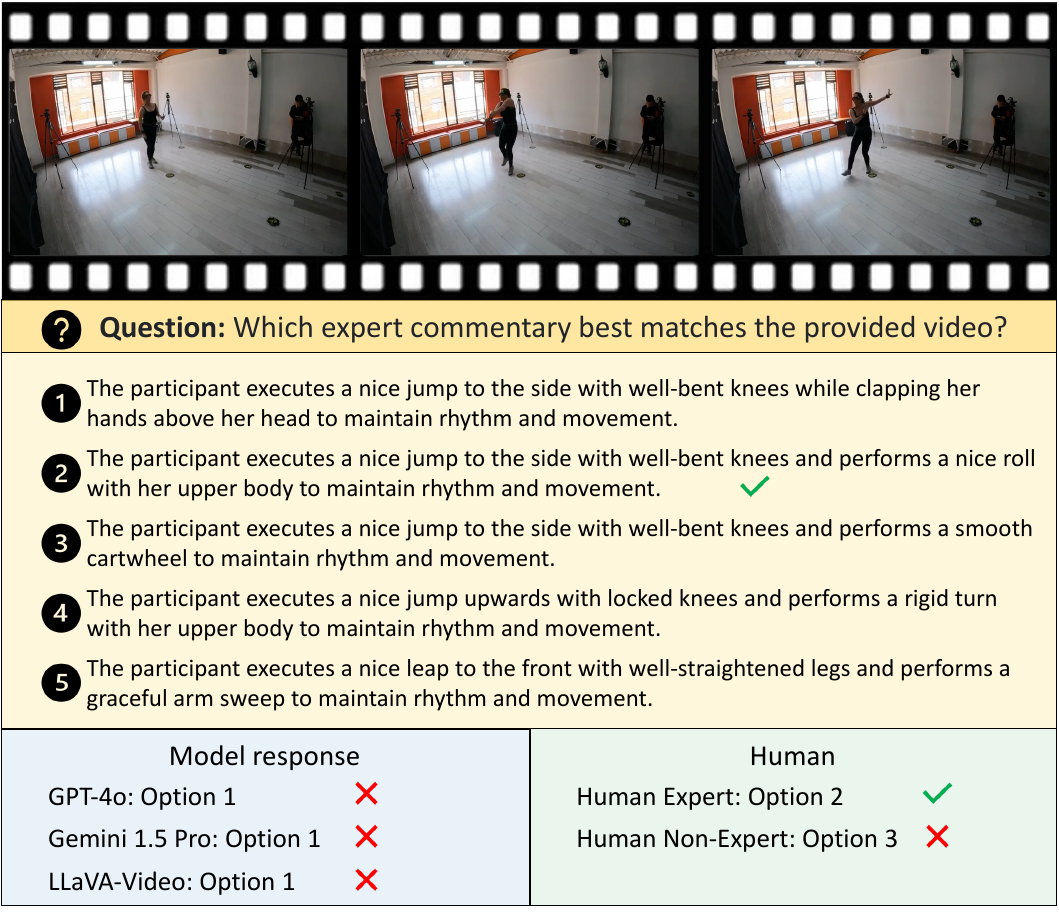}
\end{center}
% \vspace{-.15in}
   \caption{Sample 4 (Dance): None of the models select the correct answer. Only human experts identify the correct response.
}
\label{fig:sample 4}
\end{figure}

In this section, we present four QA examples (Figures~\ref{fig:sample 1}–\ref{fig:sample 4}) that range from easy to difficult. 

\textbf{Sample 1} (Figure~\ref{fig:sample 1}) represents a relatively easy example. All models, as well as both human experts and non-experts, correctly identify the correct answer. 

\textbf{Sample 2} (Figure~\ref{fig:sample 2}) presents a moderately challenging sample. Only human experts and some of the best models identify the correct answer.

\textbf{Sample 3} (Figure~\ref{fig:sample 3}) presents a more difficult case. Only human experts and GPT-4o select the correct answer.

\textbf{Sample 4} (Figure~\ref{fig:sample 4}) illustrates a particularly difficult case. None of the models are able to select the correct answer. Only human experts succeed. This highlights a significant gap between current VLM capabilities and expert-level understanding, especially for tasks that require nuanced, domain-specific reasoning.

These examples collectively highlight the limitations of current VLMs in expert-level understanding of physical human skills.

\clearpage

\section{Evaluated Capabilities in \name}
\label{sec:reasoning}

This section details the reasoning capabilities implicitly evaluated by our QA framework. 
Although each question follows a simple multiple-choice format, correctly answering them requires diverse capabilities that extend beyond visual recognition. The evaluated reasoning types include:

\begin{itemize}[itemsep=2pt, topsep=2pt, leftmargin=12pt]
    \item \textbf{Fine-grained recognition:} Detecting subtle differences in body positioning, movement patterns, and technique execution (e.g., “plays with a very tight wrist” vs. “uses a hybrid picking technique”).
    \item \textbf{Temporal reasoning:} Understanding timing precision, action phases, and temporal dependencies (e.g., “releases the ball too early” vs. “too late after the peak jump”).
    \item \textbf{Causal understanding:} Linking technique flaws to performance outcomes (e.g., “doesn’t bend knees...reduces jump height” vs. “doesn’t bend knees...prevents ball spin”).
    \item \textbf{Procedural reasoning:} Evaluating adherence to task sequences (e.g., “checks pulse before calling for help” vs. “delays calling for help and requesting an AED”).
    \item \textbf{Spatial \& geometric reasoning:} Understanding angles, trajectories, and spatial relationships (e.g., “flattens the arc on a left-handed layup” vs. “lowers the entry angle to avoid rotation”).
    \item \textbf{Domain-specific expertise:} Applying professional knowledge across sports, music, healthcare, cooking, bike repair, and dance (e.g., “maintains CPR rate at 100–120 bpm” vs. “120–140 bpm”).
    \item \textbf{Physical understanding:} Reasoning about biomechanics, force generation, and energy transfer (e.g., “angles knees inward for power” vs. “opens stance toward the rim”).
\end{itemize}

\section{Sample statistics after each filtering stage}
Starting from over 400k expert commentaries in Ego-Exo4D, Stage I reduced the number of samples to 108k ($\approx$\,--73\%). Stage III’s length similarity filtering further reduced it from 108k to 60k ($\approx$\,--44\%). The blind-LLM filtering further shrank the sample size from 60k to 4.5k ($\approx$\,--92.5\%). Finally, Stage IV expert verification led to 3.5k samples ($\approx$\,--22\% of the 4.5k), leaving about 0.9\% of the original samples. These statistics demonstrate that Stage III is critical as it filters low-quality or easily “cheatable” samples before sending them to the experts for final verification.

\end{document}